\documentclass[11pt,a4paper]{article}

\usepackage[final]{acl}

\usepackage[table]{xcolor} 
\usepackage{times}
\usepackage{latexsym}
\usepackage[T1]{fontenc}
\usepackage[utf8]{inputenc}
\usepackage{microtype}
\usepackage{inconsolata}
\usepackage{enumitem}
\usepackage{float} 
\usepackage{graphicx}
\usepackage{booktabs}
\usepackage{multirow}
\usepackage{tabularx}
\usepackage{makecell}
\usepackage{caption}
\usepackage{dblfloatfix} 
\usepackage{tcolorbox} 
\usepackage{pdfpages}
\usepackage{hyperref}
\usepackage{xurl}

\usepackage{pgfplots}    
\pgfplotsset{compat=1.18}
\usepackage{tikz}
\usetikzlibrary{shapes.geometric, shadows, positioning, arrows.meta, backgrounds, fit, calc}

\definecolor{colNorth}{RGB}{94, 235, 226}
\definecolor{colSouth}{RGB}{63, 176, 201}
\definecolor{colEast}{RGB}{45, 127, 184}
\definecolor{colWest}{RGB}{52, 85, 139}
\definecolor{colNE}{RGB}{136, 78, 160}
\definecolor{colCentral}{RGB}{192, 57, 139}
\definecolor{data_blue_light}{HTML}{E1EFF6}
\definecolor{data_blue_dark}{HTML}{4F96C4}
\definecolor{db_blue}{HTML}{8EBADC}
\definecolor{process_blue}{HTML}{93C4E0}
\definecolor{annotation_blue}{HTML}{5D9CC9}
\definecolor{analysis_blue}{HTML}{386487}
\definecolor{pink_box}{HTML}{F7B6D2}
\definecolor{pink_border}{HTML}{D6608C}
\definecolor{yellow_box}{HTML}{FFFBC8}
\definecolor{yellow_border}{HTML}{F9E754}
\definecolor{green_box}{HTML}{CDEBC5}
\definecolor{green_border}{HTML}{7DBE4E}
\definecolor{arrow_purple}{HTML}{440055}

\title{IndRegBias: A Dataset for Studying Indian Regional Biases in English and Code-Mixed Social Media Comments}

\author{
  Debasmita Panda\textsuperscript{1},
  Akash Anil\textsuperscript{1},
  Neelesh Kumar Shukla\textsuperscript{2}\\
  \textsuperscript{1}Department of Data Science and Engineering, Indian Institute of Science Education and Research, Bhopal \\
  \textsuperscript{2}Oracle Industries AI, Oracle Corporation \\
  \small{\texttt{\{debasmitap21, anila\}@iiserb.ac.in, neelesh.kumar.shukla@oracle.com}}
}
\begin{document}
\maketitle
\begin{abstract}
{\color{red} Warning: This paper consists of examples representing regional biases in Indian regions that might be offensive towards a particular region.}

While social biases corresponding to gender, race, socio-economic conditions, etc., have been extensively studied in the major applications of Natural Language Processing (NLP), biases corresponding to regions have garnered less attention. This is mainly because of  (i) difficulty in the extraction of regional bias datasets, (ii) disagreements in annotation due to inherent human biases, and (iii) regional biases being studied in combination with other types of social biases and often being underrepresented. This paper focuses on creating a dataset \texttt{IndRegBias}, consisting of regional biases in an Indian context reflected in users' comments on popular social media platforms, namely Reddit and YouTube. We carefully selected 25000 comments appearing on various threads in Reddit and videos on YouTube discussing trending topics on regional issues in India. Furthermore, we propose a multilevel annotation strategy to annotate the comments describing the severity of regional biased statements. To detect the presence of regional bias and its severity in \texttt{IndRegBias}, we evaluate open-source Large Language Models (LLM) and Indic Language Models (ILM) by using zero-shot, few-shot, and fine-tuning strategies. We observe that zero-shot and few-shot approaches show lower accuracy in detecting regional biases and severity in the majority of the LLMs and ILMs. However, the fine-tuning approach significantly enhances the performance of the LLM in detecting Indian regional bias along with its severity.  
 \end{abstract}

\section{Introduction}
\label{sec:introduction}
% Biases in NLP research and models
Natural Language Processing (NLP) has shown an impressive footprint in various applications such as Question-Answering~\cite{devlin2019bert}, Content Summarization~\cite{see2017get}, Search Engines~\cite{mitra2018introduction}, etc. NLP-based models use a large number of real-world texts in modelling the underlying task~\cite{bender2021dangers}. Consequently, NLP models are subject to various biases related to social~\cite{sap2019social}, geographic~\cite{faisal-anastasopoulos-2023-geographic}, linguistic~\cite{fleisig-etal-2024-linguistic}, and cultural~\cite{hershcovich2022challenges} factors, which may result in biased predictions~\cite{zhao2019gender}. Thus, in the recent past, there has been a surge in studying various biases~\cite{bolukbasi2016man, dixon2018measuring, sun2019mitigating} for popular NLP-based models such as Large Language Models (LLMs).

The majority of the biases, such as linguistics and culture, appear due to either over-representation or under-representation of data~\cite{hovy2016social}. However, social bias can be evident due to common societal stereotypes regarding race~\cite{dixon2018measuring}, gender~\cite{bolukbasi2016man}, economics, and region~\cite{jardim2022economic}. Although social biases based on gender, race, and economics have been studied extensively~\cite{sun2019mitigating}, only a few studies consider social biases corresponding to regions~\cite{bhatt2022recontextualizing, jha2023seegull}. However, regional stereotypes are prevalent in many diverse countries, including India, as recorded from multiple news media\footnote{(1)\href{https://www.thehindu.com/news/national/other-states/northeast-citizens-faced-racial-discrimination-amid-covid-19-outbreak-says-govt-study/article34303162.ece} {Northeast citizens faced racial discrimination amid COVID-19 outbreak} ; (2) \href{https://www.newindianexpress.com/magazine/2022/Mar/26/thegreatindian-bias-2433854.html}{The Great Indian Bias} ; (3) \href{https://economictimes.indiatimes.com/magazines/panache/bengaluru-man-demands-north-indian-be-kept-out-of-the-city-netizens-ask-him-to-tweet-in-kannada/articleshow/117554260.cms?from=mdr}{Bengaluru man demands North Indian be kept out of the city}}. Thus, this work focuses on revisiting social bias due to regional stereotypes, referred to as Regional bias (RB) in the subsequent part of the paper.

% --- TABLE 1: EXAMPLES (Wide) ---
\begin{table*}[t]
\centering
\small
\renewcommand{\arraystretch}{1.2}
\begin{tabularx}{\textwidth}{p{2cm}|X|X}
\toprule
\textbf{Language(s)} & \textbf{Comment} & \textbf{Translation} \\
\midrule
English & No offence, but Kannadigas really need to chill about language. It’s not that deep. & No offence, but Kannadigas really need to chill about language. It’s not that deep. \\
\midrule
Hindi & Arre Bihari log ka logic sunke hi dimaag hil jaata hai 'Humko sab aata hai' waale IAS ban jaate hain kaise bhai? & Just hearing Bihari people's logic shakes my brain. How do these 'we know everything' types become IAS officers? \\
\midrule
Marathi, English, Hindi & Marathi manoos ka ek hi kaam – complain about outsiders and vote for MNS. Kuch kaam bhi kar lo bhai. & The only thing a Marathi person does is complain about outsiders and vote for MNS. Do some actual work too, bro. \\
\bottomrule
\end{tabularx}
\caption{Examples of Indian regional bias expressed through different languages and sarcasm.}
\label{tab:rb_examples}
\end{table*}

\begin{figure*}[t]
    \centering
    % Reduced width to 85% of the text width to save vertical space
    \includegraphics[width=0.85\textwidth]{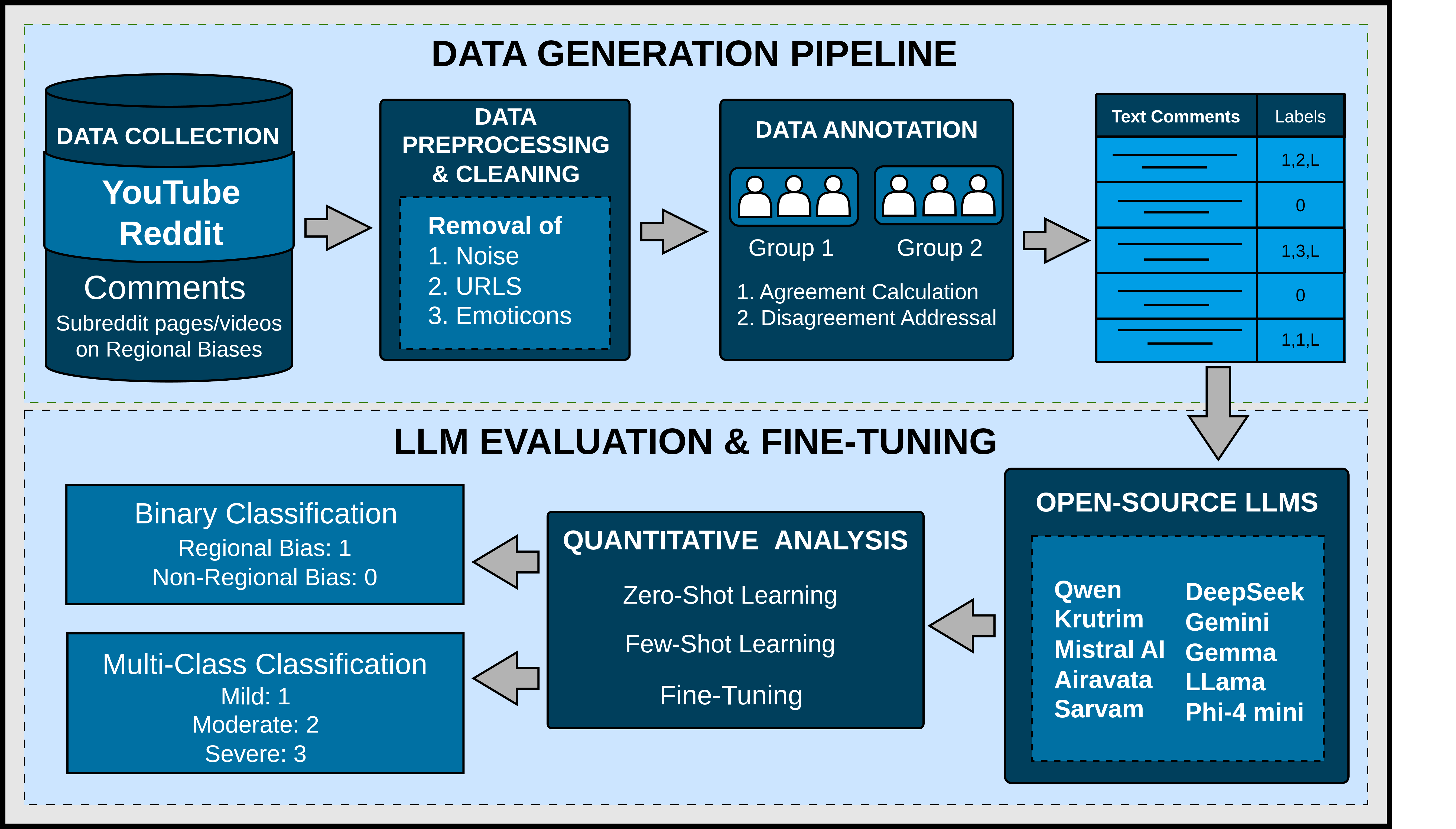}
    
    % Slightly tightened caption to save more space
    \caption{Construction of \texttt{IndRegBias} and detecting regional bias with severity using Large Language Models and Indic Language Models. (In the annotation phase, if a comment is regional bias, then the second and third labels are determined. Otherwise, the comment gets a single label of 0. For example, after the data annotation phase, a text/comment is assigned labels as (1, 2, L) where 1 is for stating the comment is regional bias, 2 states the bias is Moderate, and L states the region being targeted.}
    \label{fig:overall_project}
\end{figure*}

Regional bias broadly refers to cognitive prejudices or stereotypical assumptions directed toward individuals based on their regional identity. Table~\ref{tab:rb_examples} presents comments from popular social media, namely Reddit\footnote{https://www.reddit.com/} and YouTube\footnote{https://www.youtube.com/} by users towards Indian states or regions. It is evident that these texts carry stereotypes about natives of different regions and thus promote regional bias. Previous works~\cite{bhatt2022recontextualizing, jha2023seegull} focused on benchmarking RBs written in English. However, it is evident from Table~\ref{tab:rb_examples} that detecting RB in NLP models requires semantic understanding of the complete text, often written in transliterated and code-mixing of multiple languages. Thus, this paper focuses on creating a novel text dataset capturing regional biases in India, incorporating 36 regional boundaries\footnote{Defined by the States Reorganization Act and subsequent amendments. } referred to as state, capital territory, and union territory possessing diverse regional characteristics~\cite{census2011india}. 

% How it is difficult to get real-world regional bias dataset
% Why it makes sense to use the comments from social media?
% How do we prepare the dataset?
% How do we Annotate?
% Statistics of the benchmark
To construct the dataset of regional bias in the Indian context \texttt{IndRegBias}, this paper exploits comments from Reddit and YouTube. We chose social media comments because, due to anonymity and less regularization on social media, users tend to express their real feelings as comments corresponding to other regions~\cite{suler2004online}. Figure~\ref{fig:overall_project} presents the data generation and evaluation pipeline. In the data generation phase, we selected some of the recent topics on regional tensions in India and scraped 25000 user comments expressing their sentiments and beliefs on these topics over Reddit and YouTube. Two groups, each of three university students having diverse backgrounds and knowledge of at least three Indian languages, meticulously annotated the comments using a multilevel annotation framework resulting in three labels for a comment, i.e., given a comment $\mathcal{C}$, the first label states whether $\mathcal{C}$ is a regional bias or not ?, the second label gives the severity of $\mathcal{C}$ (Mild: 1, Moderate: 2, and Severe: 3) given $\mathcal{C}$ is a regional bias, and the third label states which region (state or union/capital territory) is being targeted given $\mathcal{C}$ is a regional bias. % We recorded high agreement scores on the annotations between the two groups. Furthermore, we resolve the disagreement using neutral annotators and a comprehensive discussion.

% Why to evaluate using Large Language Models?
% How do we evaluate using LLMs, Binary classification and Multi-class classification?
Recently, LLMs such as GPT~\cite{achiam2023gpt}, LLaMa~\cite{touvron2023llama}, and DeepSeek~\cite{liu2024deepseek} have emerged as efficient frameworks for carrying out multiple applications, e.g., question-answering and recommender systems. These frameworks are used by millions of users daily for their day-to-day queries. Most of these LLMs have been trained considering explicit knowledge of harmful content such as hate speech, slang, and other derogatory words, which is evident from their curated and restricted results. However, detecting regional biases in Indian texts remains a significant challenge due to the nation’s diverse cultures, languages, and beliefs ~\cite{bhatt2022recontextualizing}. While large language models (LLMs) have demonstrated strong performance across many tasks, they are not inherently attuned to the subtle and context-dependent aspects of Indian regional biases. This limitation arises because LLMs are trained on broad datasets where nuanced or underrepresented regional characteristics may not be adequately captured. To address this, as depicted in the LLM evaluation pipeline in Figure ~\ref{fig:overall_project}, we systematically evaluated LLMs using zero-shot, few-shot, and fine-tuning experimental setups. These settings enable us to assess how well LLMs can generalize to the task of regional bias detection (binary classification) and severity ranking (multiclass classification), and to what extent their performance improves with domain-specific examples or further adaptation.

The rest of the paper is organized as follows: Section~\ref{sec:related_studies} presents the related studies on bias and NLP. Section~\ref{sec:data_creation} describes the proposed regional biased dataset in the Indian context, followed by Section~\ref{sec:experiments} presenting the experimental setups for LLM evaluation. Section~\ref{sec:results_discussion} presents the results and discussion. Finally, Section~\ref{sec:conclusion} concludes the paper. 

\section{Related Studies}
\label{sec:related_studies}
\subsection*{Bias in NLP and Regional Bias}
Social biases in NLP models have primarily been studied in the context of gender~\cite{bolukbasi2016man, sun2019mitigating}, race~\cite{dixon2018measuring}, and religion~\cite{abid2021persistent}. These works have demonstrated that a majority of the language models encode the inherent stereotypes and may result in biased prediction~\cite{bender2021dangers}. While regional identity is an important factor under the social system, less attention has been paid to studying and curating resources corresponding to regional biases~\cite{Hovy2021FiveSO}. Recently,~\cite{adebayo2024regional} discusses the regional biases in monolingual English language models and contributes an investigation framework specifically tailored for low-resource regions, revealing that there are multiple geographic variations in the word embeddings by BERT~\cite{devlin2019bert} while they were assumed to be similar. Further,~\cite{borah2025towards} identifies topical differences in gender bias across different regions and proposes a region-aware bottom-up approach for bias assessment. To measure the hierarchical regional bias,~\cite{li2022herb} proposes a metric \texttt{HERB} that utilizes the information from the sub-region clusters to quantify the bias in pre-trained language models. 

While these studies consider regional aspects, they are limited to analytics supported by other types of biases.

\subsection*{Regional Bias in Multilingual Indian Regions}
Regional biases in Indian regions are commonly prevalent and are often seen in terms of dialect variations~\cite{kumar2021dialect}, transliteration~\cite{roark2012transliteration}, and code-mixed uses~\cite{bali2014code}. In a recent work,~\cite{bhatt2022recontextualizing} released a fairness evaluation corpus covering stereotypes pertaining to region and religion axes relevant to the Indian context in the English language. Further,~\cite{jha2023seegull} introduces a novel benchmark \texttt{SeeGULL} in English containing stereotypes about identity groups spanning 178 countries across 8 different geopolitical regions across 6 continents, as well as state-level identities within the US and India using generative models. While these resources are related to Indian regions, they have been curated in English, which might lose the inherent bias characteristics of the local regions within India. 

Although not specific to regional bias,~\cite{sahoo2024indibias} introduces \texttt{IndiBias}, a benchmark dataset designed particularly for evaluating social biases in the Indian context and capturing different variants of social biases in the Hindi and English languages. 

These datasets bridge the gap in social biases with Indian contexts, but as discussed, regional biases in real-world communication are more nuanced and often found transliterated or code-mixed. Thus, our paper focuses on introducing \texttt{IndRegBias}, consisting of real-world discussions (using comments from social media) over regional issues, which can bridge the gap for regional bias datasets in the Indian context.

%Social bias and stereotype detection using natural language processing (NLP) and LLM have seen a rise in recent time~\cite{mozafari2020hate, viswanath2023fairpy, guo2024bias, liang2021towards}. However, very few studies exclusively consider the regional biases that are prevalent in countries such as India with large geographical boundaries and the largest population having different native languages and cultures~\cite{li-etal-2022-herb}. 

%Traditional NLP models often fail at capturing the regional biases that are code-mixed, culture-driven, sarcasm-embedded, and consisting of emoticons for humor~\cite{singh2024predicting, sasidhar2020emotion}. Recently, transformer-based NLP models such as IndicBERT~\cite{kakwani2020indicnlpsuite}, GPT-4~\cite{achiam2023gpt}, and MuRIL~\cite{khanuja2021muril} have shown the potential of LLMs in capturing these aspects and performing efficiently in zero-shot and few-shot settings. However, due to the unavailability of focused labelled data for regional biases, these models can learn to detect other inherent societal biases and might be inefficient for local regional bias statements~\cite{li-etal-2022-herb}. Therefore, this paper attempts to bridge this gap by contributing a novel dataset for regional biases prevalent in Indian social interactions. 

\section{Creation Of \texttt{IndRegBias}}
\label{sec:data_creation}
The creation of \texttt{IndRegBias} is motivated by the scarcity of Indian benchmark datasets, particularly for regional biases. While the other benchmarks created in~\cite{sahoo2024indibias, jha-etal-2023-seegull} capture Indian contexts, they are based on translation primarily from the English language. However, with the freedom of speech on social media, a majority of users post their stereotypes anonymously~\cite{suler2004online}. We argue that NLP-based models should be robustly trained over such kinds of texts to promote regional inclusivity.  

\subsection{Data Collection}
To prepare regional bias datasets over social media, we consider two popular online discussion platforms, namely, Reddit and YouTube. We selected subreddit pages focusing on discussions about the stereotypes and biases targeting particular regions and states in India. Similarly, we selected YouTube channels that specifically discuss individual states, the biases related to them, the prevalent problems in the state, the issues associated with the identity of the people, and interviews with residents.

We extracted the comments using \texttt{praw} (Python Reddit API Wrapper) \cite{praw_documentation} for Reddit and \texttt{googleapiclient} \cite{google_api_client} for YouTube. The comments collected from these platforms contain numerous spam comments that are irrelevant to our work. Thus, after cleaning and preprocessing, a final dataset of 25,000 comments was curated for annotation. Further, as the comments were collected from videos or subreddit pages belonging to different Indian regions, where the languages are mixed, such as English, Hinglish (a mix of Hindi and English), a mix of Bengali and English, a mix of Malayalam and English, or Marathi and English, etc., it results in a transliterated and code-mixed dataset naturally representing the Indian diversity.

\subsection{Data Annotation}
\label{subsec:data annotation}
The nuances in social media comments, particularly in the Indian context, make the annotation task very challenging. Thus, we propose a multilevel annotation policy to ensure high-quality labeling along with maintaining consistency. 

\subsubsection{Multilevel Annotation Policy}
\label{sec:annotation_policy}
The goal of the annotation task was to categorize social media comments into regional biases or non-regional biased categories. The annotation process was divided into three sub-tasks:

\textbf{Task 1: Bias Identification}\\
Each comment was classified into one of two categories:
\begin{itemize}
    \item \textbf{Regional Bias (RB: 1):} The comment propagates a stereotype (positive or negative) regarding a state, region, caste, politics, or culture associated with a specific Indian region.
    \item \textbf{Non-Regional Bias (NRB: 0):} The comment does not exhibit stereotypes or is unrelated to the objective.
\end{itemize}

\textbf{Task 2: Severity Scoring}
The comments classified as RB in Task 1 were assigned a severity score based on the definitions in Table \ref{tab:severity_levels}. Positive stereotypes were explicitly mapped to the ``Mild'' category.

\begin{table}[h]
    \centering
    \small
    % Use tabularx to fit to column width. 
    % The 'X' column automatically calculates width and wraps text.
    \begin{tabularx}{\columnwidth}{@{}c l X@{}} 
    \toprule
    \textbf{Score} & \textbf{Level} & \textbf{Definition} \\
    \midrule
    0 & None & No regional bias present. \\
    1 & Mild & Slight bias, satire, or \textbf{positive stereotypes} (e.g., ``Himachalis are very simple''). \\
    2 & Moderate & Clear bias reinforcing negative stereotypes with intent to mock. \\
    3 & Severe & Strong bias promoting discrimination, racism, or dehumanization. \\
    \bottomrule
    \end{tabularx}
    \caption{Severity scoring criteria for regional bias.}
    \label{tab:severity_levels}
\end{table}

\textbf{Task 3: Target Region Assignment}
Annotators identified the target(s) of the comment from a predefined list of regions (e.g., \textit{North-India, NorthEast-India, Central India, etc.}) and 30+ specific States/Union Territories (e.g., \textit{Punjab, Tamil Nadu, etc.}). 

The data annotation involved two groups, each with three university students with diverse backgrounds and understanding of at least three Indian languages, including English. Both of the groups annotated the same set of data using the above annotation policy. To validate the reliability of our annotation process, we employed Cohen's Kappa ($\kappa$)~\cite{cohen1960coefficient} coefficient for calculating the agreement score. We selected this metric over simple percent agreement because it explicitly accounts for the possibility of the agreement occurring by chance, providing a more robust measure of inter-annotator reliability for categorical items. The Cohen's Kappa ($\kappa$) coefficient is calculated using the following equation:

\begin{equation}
    \kappa = \frac{p_o - p_e}{1 - p_e}
\end{equation}

\noindent where $p_o$ represents the relative observed agreement among raters, and $p_e$ represents the hypothetical probability of chance agreement. Upon calculation, the Cohen's kappa value for binary classification (RB vs NRB) came to be \textbf{0.91}, and \textbf{0.83} for multi-class classification (severity and region assignment combined), indicating strong agreement. Further disagreements were resolved through comprehensive discussions among the annotators, including one neutral annotator. Table \ref{tab:annotation_examples} demonstrates a sample of annotated \texttt{IndRegBias}.

\begin{table*}[t] % Changed [h] to [t] for better stability in 2-column mode
    \centering
    \small
    \renewcommand{\arraystretch}{1.2} % Adds a little breathing room to rows
    % X column type automatically fills the remaining width
    \begin{tabularx}{\textwidth}{>{\raggedright\arraybackslash}X c c l} 
    \toprule
    \textbf{Comment} & \textbf{Is RB?} & \textbf{Severity} & \textbf{state/region} \\
    \midrule
    ``Biharis have a weird accent.'' & 1 & 1 & Bihar \\
    ``Jharkhand is just a jungle.'' & 1 & 2 & Jharkhand \\
    ``North Indians are just casteist and illiterate'' & 1 & 3 & North-India \\
    ``Himachalis are so sweet.'' & 1 & 1 & Himachal Pradesh \\
     ``Proud to be an Assamese.'' & 0 & 0 & NA \\
    \bottomrule
    \end{tabularx}
    \caption{A sample of \texttt{IndRegBias} with fine-grained annotation using the multi-level annotation policy}
    \label{tab:annotation_examples}
\end{table*}

\iffalse
% --- TABLE 2: ANNOTATION SNAPSHOT (Wide) ---
\begin{table*}[!htbp] 
\centering
\small
\renewcommand{\arraystretch}{1.2}
\begin{tabularx}{\textwidth}{X c c l}
\toprule
\textbf{Comment} & \textbf{is RB?} & \textbf{severity} & \textbf{state} \\ 
\midrule
Punjabis - loud, dumb, flashy and drunk.                                  & 1                & 2                & Punjab           \\
Biharis: Either illiterate or will murder you. sometimes both. & 1                & 3                & Bihar            \\
uttar pradesh- communal riots, gangsters, katta                  & 1                & 3                & Uttar Pradesh    \\
Haryana people are backward.                                   & 1                & 2                & Haryana          \\ 
\bottomrule
\end{tabularx}
\caption{Snapshot of the Indian Regional Bias dataset with fine-grained annotation using the Multi-level annotation framework discussed above.}
\label{tab:data_snap_fullwidth_st}
\end{table*}
\fi

\subsection{Data Analysis}
\label{subsec:data_analysis}
After annotation of 25000 comments, \texttt{IndRegBias} consists of 13,015 (52.1\%) regional biases, while 11,985 (47.9\%) are non-regional biases. Under the severity label for regional biased comments approximately 29\% are Mild, 51\% are Moderate, and 20\% are Severe. 

\subsubsection{Data Distribution based on Region}

Figure~\ref{fig:region_dist_bar} presents the distribution of regional biases \texttt{IndRegBias} categorized into 6 Indian regions (with overlapping possible if a comment targets more than one region), namely North, South, East, West, NorthEast, and Central (based on collections of some of the states that are closer geographically).  We observe that approximately 40\% of biased comments target South Indians followed by North India with 26\%. Further, East and West Indian regions are targeted by approximately 16\% and 10\% of comments, respectively. We record a lower number of comments, approximately 5\% and 2\% of comments targeting NorthEast and Central region respectively.  

\subsubsection{Data Distribution based on States/Territories}

\begin{figure}[!t]
    \centering
    \begin{tikzpicture}
        \begin{axis}[
            % --- Chart Type ---
            xbar,
            width=0.95\columnwidth, 
            height=12cm, 
            bar width=6pt,
            % --- Axes Setup ---
            xmin=0, xmax=2400, % Increased max to fit larger number fonts
            xlabel={Number of Comments},
            ylabel={},
            % --- Labels and Ticks ---
            ytick={0,1,...,28}, 
            yticklabels={
                Tripura, Meghalaya, Manipur, Sikkim, Mizoram, 
                Arunachal Pr., Assam, Uttarakhand, Andhra Pr., 
                Nagaland, Telangana, Himachal Pr., Odisha, 
                Chhattisgarh, Madhya Pr., Jharkhand, Haryana, Rajasthan, 
                J\&K, Punjab, Maharashtra, Uttar Pradesh, 
                Tamil Nadu, Gujarat, Bihar, Karnataka, 
                West Bengal, Goa, Kerala
            },
            yticklabel style={font=\small},
            % --- Values on Bars ---
            nodes near coords,
            nodes near coords style={font=\footnotesize, color=black, anchor=west},
            % --- Aesthetics ---
            grid=major,
            grid style={dashed, gray!30},
            axis x line=bottom,
            axis y line=left,
            enlarge y limits=0.02
        ]
            % --- Data ---
            \addplot[fill=teal!70, draw=black] coordinates {
                (12,0) (16,1) (27,2) (27,3) (30,4) 
                (32,5) (49,6) (65,7) (72,8) (98,9) 
                (106,10) (107,11) (111,12) (163,13) (173,14) 
                (210,15) (217,16) (218,17) (218,18) (412,19) 
                (438,20) (599,21) (614,22) (832,23) (871,24) 
                (887,25) (1091,26) (1095,27) (1857,28)
            };
        \end{axis}
    \end{tikzpicture}
    \caption{Distribution of regional biased comments across Indian states.}
    \label{fig:state_dist_bar}
\end{figure}
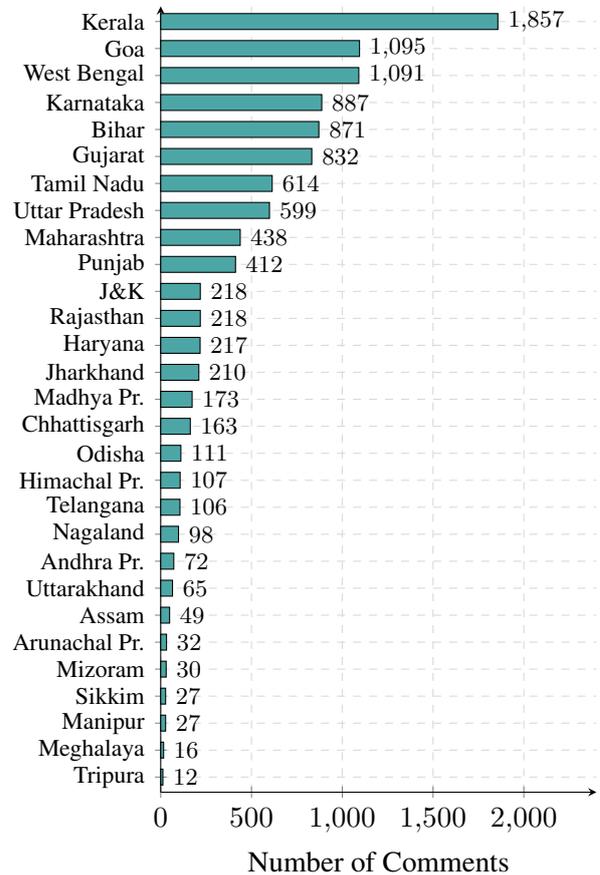

Figure~\ref{fig:state_dist_bar} shows the distribution of regional biases categorized in 29 Indian states. We observe that the distribution loosely describes the long-tailed distribution where few of the states, such as Kerala, Goa, West Bengal, Karnataka, Bihar, and Gujarat, are targeted approximately 60\% of the time. 

\begin{figure}[!t]
    \centering
    \begin{tikzpicture}
        \begin{axis}[
            % --- Chart Type ---
            xbar,
            width=0.95\columnwidth, 
            height=6cm, % Reduced height to fit 6 bars compactly
            bar width=15pt, % Thicker bars for better visibility
            % --- Axes Setup ---
            xmin=0, 
            xmax=7000, % Fixed: Increased to accommodate the max value (5820)
            xlabel={Number of Comments},
            ylabel={},
            % --- Labels and Ticks ---
            ytick={0,1,2,3,4,5}, 
            yticklabels={
                Central-India, NorthEast-India, West-India, East-India, North-India, South-India
            },
            yticklabel style={font=\small},
            % --- Values on Bars ---
            nodes near coords,
            nodes near coords style={font=\footnotesize, color=black, anchor=west},
            % --- Aesthetics ---
            grid=major,
            grid style={dashed, gray!30},
            axis x line=bottom,
            axis y line=left,
            enlarge y limits=0.15 % Adds spacing above/below first/last bars
        ]
            % --- Data ---
            \addplot[fill=teal!70, draw=black] coordinates {
                (5820,5) (3844,4) (2307,3) (1508,2) (827,1) (339,0)
            };
        \end{axis}
    \end{tikzpicture}
    \caption{Distribution of data across different regions in India.}
    \label{fig:region_dist_bar}
\end{figure}
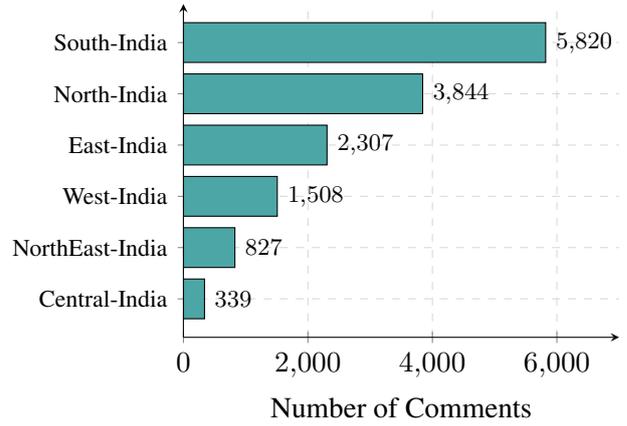

\subsection{Qualitative Analysis of Regional Biased Comments}
\label{subsection:quali_analysis}

% --- TABLE: STEREOTYPES ---
\begin{table*}[!t]
    \centering
    \small
    \renewcommand{\arraystretch}{1.3}
    \begin{tabularx}{\textwidth}{l p{5cm} X}
        \toprule
        \textbf{Theme} & \textbf{States/Regions} & \textbf{Stereotypes} \\
        \midrule
        \textbf{Crime \& Corruption} & 
        Delhi, Uttar Pradesh, Tripura, J\&K, Goa & 
        Frequently labelled with stereotypes related to high crime rates, corruption, drug trafficking, and organised crime. \\
        \midrule
        \textbf{Social Intolerance} & 
        Delhi, Mizoram, Karnataka, Haryana, Uttar Pradesh, J\&K, Manipur & 
        Stereotypes concerning racism, arrogance, misogyny, casteism, and anti-national sentiments. \\
        \midrule
        \textbf{Socio-Economic Failure} & 
        Bihar, Uttar Pradesh, Jharkhand & 
        Branded as ``failed'' or ``backward,'' focusing on poverty, unemployment, illiteracy, and caste conflict (framed as a ``national burden''). \\
        \bottomrule
    \end{tabularx}
    \caption{Categorization of severe regional bias comments}
    \label{tab:stereotypes}
\end{table*}

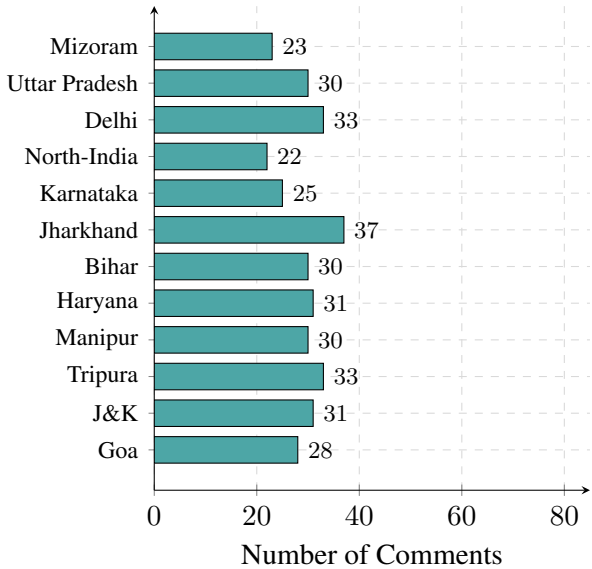
\begin{figure}[!t]
    \centering
    \begin{tikzpicture}
        \begin{axis}[
            % --- Chart Type ---
            xbar,
            width=0.95\columnwidth, 
            height=8cm, % Reduced height slightly as there are fewer bars (9 vs 29)
            bar width=10pt, % Increased width for better visibility
            % --- Axes Setup ---
            xmin=0, xmax=85, % FIXED: Reduced from 2400 to 85 to fit the data range (Max value is 72)
            xlabel={Number of Comments},
            ylabel={},
            % --- Labels and Ticks ---
            ytick={0,1,2,3,4,5,6,7,8,9,10,11}, 
            yticklabels={Goa, J\&K, Tripura, Manipur, Haryana, Bihar, Jharkhand, Karnataka, North-India, Delhi, Uttar Pradesh, Mizoram},
            yticklabel style={font=\small},
            % --- Values on Bars ---
            nodes near coords,
            nodes near coords style={font=\footnotesize, color=black, anchor=west},
            % --- Aesthetics ---
            grid=major,
            grid style={dashed, gray!30},
            axis x line=bottom,
            axis y line=left,
            enlarge y limits=0.1
        ]
            % --- Data ---
            % FIXED: Added missing '(' at the start of the coordinates
            \addplot[fill=teal!70, draw=black] coordinates {
                (28,0) (31,1) (33,2) (30,3) (31,4) (30,5) (37,6) (25,7) (22,8) (33,9) (30,10) (23,11)
            };
        \end{axis}
    \end{tikzpicture}
    \caption{Distribution of severe bias comments across different states.}
    \label{fig:state_dist_bar_severe}
\end{figure}

We perform qualitative analysis of \texttt{IndRegBias}, which resulted in distinct patterns for the severity of the biases present in the comments, having aggression and using derogatory language frequently to describe their ideas over the states or the people of the state. We categorized the states having severe regional biases into three main themes, namely Crime and Corruption, Social Intolerance, and Socio-Economic Failures, as summarized in Table \ref{tab:stereotypes}. Figure~\ref{fig:state_dist_bar_severe} shows the distribution of severe comments under the three themes. We observed that these three themes share approximately a similar number of comments when distributed qualitatively. For example, Delhi received 33 comments under Crime and Corruption and Social Intolerance, and Jharkhand and Bihar received 37 and 30, respectively.

%\subsubsection{Representation of Low-Resource States}
%Fig \ref{fig:state_dist_bar} highlights states with significantly fewer comments. This disparity is attributed to geographic/digital barriers and marginalisation in national discourse.

\section{Experimental Setups and Analysis}
\label{sec:experiments}
%\subsection{Experimental Flow}
%\label{subsec:exp_flow}

The experimental analysis begins with evaluating the models in zero-shot and few-shot settings, followed by fine-tuning to improve their identification of regional biases. For zero-shot \cite{kojima2022large} and few-shot \cite{brown2020language}, we have used the chain-of-thought prompting \cite{wei2022chain} technique described in Section \ref{subsec:system_prompts} to classify comments into binary or multi-class labels.In the few-shot setting, we have conducted experiments across different support settings, of which the support was selected randomly, and inferencing was conducted on the rest of the comments. The rationale was provided in Section \ref{subsec:Rationale_support_set}. 
For the fine-tuning experiment, we have employed Parameter-Efficient Fine-Tuning (PEFT) \cite{xu2023parameter} using Low-Rank Adaptation (LoRA) \cite{hu2021lora}. This approach freezes the pre-trained weights and injects trainable adapters, significantly reducing computational overhead. We have used a random dataset split of \textbf{70\% Train, 10\% Validation, and 20\% Test}. To ensure robust results and mitigate overfitting, we have implemented \textbf{5-Fold Stratified Cross-Validation}. We have applied two fine-tuning techniques: (i) \textbf{Instruction-Based Supervised Fine Tuning (SFT)} for binary classification for detecting a comment as regional bias or not, and \textbf{Classification-Based SFT} for multi-class classification, where we replace the generative layer with a classification head to predict severity levels (1--3) directly. A \texttt{WeightedRandomSampler} is employed to address class imbalance. All hyperparameters for the fine-tuning experiments are listed in Appendix Section~\ref{subsec:hyparam}.

\section{Results and Discussion}
\label{sec:results_discussion}

We conducted systematic experiments to evaluate LLMs and ILMs for the detection of regional biases in \texttt{IndRegBias} using zero-shot, few-shot, and fine-tuning approaches.\footnote{All the prompts are available in the Appendix. The dataset and codes are available on \href{https://github.com/debby123p/IndRegBias-A-Dataset-for-Studying-Indian-Regional-Biases-in-English-and-Code-Mixed-Languages}{GitHub}.} At first, detecting regional bias is achieved by employing the language models as binary classifiers. Additionally, to assess severity, we employ a multi-class classification method wherein the language models predict a severity class (i.e., Mild, Moderate, and Severe). We utilize conventional machine learning measures to evaluate the performance of different models.

%The experimental evaluation was conducted in two phases: binary classification (Regional Bias vs. Non-Regional Bias) on the complete dataset followed by multi-class classification (Severity Levels) for Regional Biases only. We use Accuracy (Acc), Precision (P), and F1-score (F1) as the primary evaluation metrics.

\subsection{Regional Bias (RB) vs Non-Regional Bias (NRB): Binary Classification}
\label{subsec:bin_class}

Table~\ref{tab:zeroshot_results} presents the performance of 11 open-source large language models (8 LLMs with variants + 3 ILMs) for identifying comments as RB or NRB using zero-shot inference\footnote{A small snapshot of IndRegBias consisting of 5000 comments was evaluated using proprietary model GPT-4o and LLama 4 Scout. We observe that open-source models such as Qwen were achieving similar performance. Table \ref{subsec:extra_exp} gives the results of this setting in the Appendix}. It is evident that Qwen3 (8B and 32 B) consistently outperformed all other LLMs. This might be due to extensive multilingual pre-training of Qwen with 119 languages, including Indian languages \cite{qwen3report}, capturing the regional semantics. Moreover, Krutrim-2 outperforms all other ILMs and maintains competitiveness in comparison to Qwen3. This may be attributed to Krutrim-2's utilization of Mistral-Nemo and its targeted pre-training in 13 Indian languages \cite{kallappa2025krutrim}. 

Inferencing the above language models in the few-shot setting is costly in terms of GPU requirements and time. Thus, we select the Qwen3-8B model from Table~\ref{tab:zeroshot_results},  as it is competitive to Qwen3-32B but requires less memory. We first do a preliminary experiment on a smaller snapshot for selecting supports in a few-shot evaluation\footnote{Refer to Section \ref{subsec:prelim_results} in Appendix}. As shown in Table \ref{tab:combined_all_experiments}, providing 50 regional bias examples (Exp-1) yielded the most improvement, boosting the F1-score to 0.82 over the zero-shot scores. The balanced support set (Exp-2) attained the highest Precision (0.85) but experienced a reduction in F1-score (0.75), signifying a notable trade-off in recall. Exp-3 improved both Precision (0.80) and F1-score (0.81).

As discussed in Section~\ref{sec:experiments}, we select the splits 70\%, 10\%, and 20\% for \texttt{IndRegBias} and evaluate the performance of fine-tuning Qwen3-8B and Qwen3-32B in a five-fold stratified cross-validation setting. Table~\ref{tab:finetune_results} presents the average of fine-tuning results compared with the zero-shot performance of the same testing examples. It is evident that fine-tuned Qwen3-8B and Qwen3-32B models consistently achieved better performance than the zero-shot setting. This justifies the requirement of a focused dataset to train the existing LLMs for capturing the regional biases. 

% --- TABLE 5: ZERO-SHOT BINARY ---
\begin{table}[t]
    \centering
    \small
    \renewcommand{\arraystretch}{1.1}
    \caption{Zero-Shot Binary Classification for Regional Bias Detection in \texttt{IndRegBias}. [Acc: Accuracy, P: Precision, F1: F1-Score]}
    \label{tab:zeroshot_results}
    \begin{tabular*}{\columnwidth}{@{\extracolsep{\fill}}l c c c}
        \toprule
        \textbf{Model} & \textbf{Acc} & \textbf{P} & \textbf{F1} \\
        \midrule
        \textbf{Qwen3-8B} & \textbf{0.74} & 0.71 & \textbf{0.78} \\
        \textbf{Qwen3-32B} & \textbf{0.74} & \textbf{0.79} & 0.72 \\
        Krutrim-2 & 0.73 & 0.68 & \textbf{0.78} \\
        Mistral-7B-v0.3 & 0.70 & 0.65 & 0.76 \\
        Gemini-2.5-Pro & 0.69 & 0.64 & 0.76 \\
        Airavata & 0.58 & 0.60 & 0.58 \\
        Sarvam-M & 0.57 & 0.63 & 0.49 \\
        Llama-3.2-3B & 0.55 & 0.57 & 0.57 \\
        DeepSeek-R1 & 0.54 & 0.53 & 0.69 \\
        Gemma-1.1-7B-IT & 0.52 & 0.52 & 0.69 \\
        Mistral-Nemo & 0.53 & 0.53 & 0.69 \\
        Phi-4-Mini & 0.48 & 0.60 & 0.00 \\
        \bottomrule
    \end{tabular*}
\end{table}

% --- TABLE 7: FEW-SHOT BINARY ---
\begin{table}[t]
    \centering
    \footnotesize
    \renewcommand{\arraystretch}{1.1}
    \setlength{\tabcolsep}{2pt}
    \caption{Few-Shot (FS) vs Zero-Shot (ZS) Performance by Qwen3-8B. [R: Regional Bias, N: Non-Regional Bias]}
    \label{tab:combined_all_experiments}
    \begin{tabular*}{\columnwidth}{@{\extracolsep{\fill}}l c c c c c c}
        \toprule
        & \multicolumn{2}{c}{\textbf{Exp-1}} & \multicolumn{2}{c}{\textbf{Exp-2}} & \multicolumn{2}{c}{\textbf{Exp-3}} \\
        & \multicolumn{2}{c}{\scriptsize{(50 R)}} & \multicolumn{2}{c}{\scriptsize{(25R/25N)}} & \multicolumn{2}{c}{\scriptsize{(30R/20N)}} \\
        \cmidrule(lr){2-3} \cmidrule(lr){4-5} \cmidrule(lr){6-7}
        \textbf{Metric} & \textbf{ZS} & \textbf{FS} & \textbf{ZS} & \textbf{FS} & \textbf{ZS} & \textbf{FS} \\
        \midrule
        Precision & 0.70 & \textbf{0.74} & 0.71 & \textbf{0.85} & 0.71 & \textbf{0.80} \\
        F1-Score  & 0.77 & \textbf{0.82} & 0.77 & 0.75 & 0.78 & \textbf{0.81} \\
        \bottomrule
    \end{tabular*}
\end{table}

%\subsubsection{Fine-Tuning Results}
%\label{subsubsec:FT_bin_results}
%The fine-tuned Qwen3-8B and Qwen3-32B models achieved comparable high performance, with the 8B model marginally outperforming the 32B model in detecting regional biases (Precision: 0.904 vs 0.902; F1: 0.902 vs 0.900) as presented in \ref{tab:finetune_results}). 

% --- TABLE 8: FINE-TUNING BINARY ---
\begin{table}[t]
    \centering
    \small 
    \renewcommand{\arraystretch}{1.1} 
    \setlength{\tabcolsep}{2pt} 
    \caption{Comparison of Fine-Tuning (FT) vs. Zero-Shot (ZS) for Qwen3 Models.}
    \label{tab:finetune_results}
    \begin{tabular*}{\columnwidth}{@{\extracolsep{\fill}}l c c c c}
        \toprule
        & \multicolumn{2}{c}{\textbf{Qwen3-8B}} & \multicolumn{2}{c}{\textbf{Qwen3-32B}} \\
        \cmidrule(lr){2-3} \cmidrule(lr){4-5}
        \textbf{Metric} & \textbf{ZS} & \textbf{FT} & \textbf{ZS} & \textbf{FT}\\
        \midrule
        Precision & 0.692 & \textbf{0.902} & 0.790 & \textbf{0.900} \\
        F1-Score  & 0.772 & \textbf{0.904} & 0.724 & \textbf{0.902} \\
        \bottomrule
    \end{tabular*}
\end{table}

\subsection{Detecting Severity of Regional Bias: Multi-Class Classification}
\label{subsec:multi_class}

We selected the four best-performing models from Table~\ref{tab:zeroshot_results} for zero-shot multi-class severity classification of the regional bias comments. Table \ref{tab:severity_results} presents the result of severity prediction using the zero-shot approach. It is evident that Mistral-7B-v0.3 achieved the highest Precision for the `Mild' and `Moderate' classes (0.64) and the best F1-score for the `Severe' class (0.45). Qwen3-8B excelled for the `Severe' class with Precision (0.44) and for the `Moderate' class with F1-score (0.66). Although Mistral-Nemo-Base performed comparably to Qwen3-8B, Qwen3-32B lagged in both the `Mild' and `Severe' categories.

% --- TABLE 9: ZERO-SHOT MULTI ---
\begin{table}[t!]
    \centering
    \footnotesize % Reduced font size
    \renewcommand{\arraystretch}{1.0} % Tighter spacing
    \setlength{\tabcolsep}{3pt}
    \caption{Zero-Shot Multi-class severity detection for Regional Biases. [P: Precision, F1: F1-Score]}
    \label{tab:severity_results}
    \begin{tabular*}{\columnwidth}{@{\extracolsep{\fill}}l c c c c c c}
        \toprule
        & \multicolumn{2}{c}{\textbf{Mild}} & \multicolumn{2}{c}{\textbf{Mod.}} & \multicolumn{2}{c}{\textbf{Severe}} \\
        \cmidrule(lr){2-3} \cmidrule(lr){4-5} \cmidrule(lr){6-7}
        \textbf{Model} & \textbf{P} & \textbf{F1} & \textbf{P} & \textbf{F1} & \textbf{P} & \textbf{F1} \\
        \midrule
        Qwen3-8B       & 0.57 & 0.57 & 0.59 & \textbf{0.66} & \textbf{0.44} & 0.22 \\
        Qwen3-32B      & 0.51 & 0.58 & 0.62 & 0.57 & 0.39 & 0.36 \\
        Mistral-7B     & \textbf{0.64} & \textbf{0.62} & \textbf{0.64} & 0.54 & 0.36 & \textbf{0.45} \\
        Mistral-Nemo   & 0.52 & 0.58 & 0.61 & 0.59 & 0.41 & 0.35 \\
        Krutrim-2      & 0.48 & 0.59 & 0.61 & 0.55 & 0.41 & 0.32 \\
        \bottomrule
    \end{tabular*}
\end{table}

For few-shot multi-class classification, we evaluate the best-performing model (Mistral-7B-v0.3) using three support configurations. As shown in Table \ref{tab:fewshot_multiclass_vs_zero}, performance relative to the zero-shot baseline is mixed. The larger balanced set (Exp B) improved stability, significantly boosting the `Moderate' F1-score (0.54 to 0.64) and `Severe' Precision (0.36 to 0.41), though at the cost of `Severe' recall.

\begin{table}[t]
    \centering
    \small
    \renewcommand{\arraystretch}{1.1}
    \setlength{\tabcolsep}{2pt}
    \caption{Few-Shot Multi-class Classification Results compared to Zero-Shot baselines.[P: Precision, F1: Score]}
    \label{tab:fewshot_multiclass_vs_zero}
    \begin{tabular*}{\columnwidth}{@{\extracolsep{\fill}}l c c c c}
        \toprule
        & \multicolumn{2}{c}{\textbf{Zero-Shot}} & \multicolumn{2}{c}{\textbf{Few-Shot}} \\
        \cmidrule(lr){2-3} \cmidrule(lr){4-5}
        \textbf{Class} & \textbf{P} & \textbf{F1} & \textbf{P} & \textbf{F1} \\
        \midrule
        \multicolumn{5}{l}{\textit{Exp A: 9 Examples (3 per class)}} \\
        \midrule
        Mild (1)     & \textbf{0.64} & \textbf{0.62} & 0.58 & \textbf{0.62} \\
        Moderate (2) & \textbf{0.64} & 0.54 & \textbf{0.64} & \textbf{0.58} \\
        Severe (3)   & 0.36 & \textbf{0.45} & \textbf{0.40} & 0.44 \\
        \midrule
        \multicolumn{5}{l}{\textit{Exp B: 30 Examples (10 per class)}} \\
        \midrule
        Mild (1)     & 0.64 & \textbf{0.62} & \textbf{0.66} & 0.61 \\
        Moderate (2) & \textbf{0.64} & 0.54 & \textbf{0.65} & \textbf{0.64} \\
        Severe (3)   & 0.36 & \textbf{0.45} & \textbf{0.41} & 0.41 \\
        \midrule
        \multicolumn{5}{l}{\textit{Exp C: 30 Examples (Imbalanced)}} \\
        \midrule
        Mild (1)     & \textbf{0.64} & 0.62 & 0.62 & \textbf{0.63} \\
        Moderate (2) & \textbf{0.64} & 0.54 & 0.62 & \textbf{0.67} \\
        Severe (3)   & 0.35 & \textbf{0.45} & \textbf{0.48} & 0.31 \\
        \bottomrule
    \end{tabular*}
\end{table}

For fine-tuning in the multi-class severity detection setting, we fine-tuned Mistral-7B-v0.3 using 5-fold stratified cross-validation on regional bias comments only. The results presented in Table \ref{tab:zeroshot_vs_finetune_severity} demonstrate significant improvements over zero-shot across all classes. Notably, the 'Severe' category has considerable gains, validating the effectiveness of fine-tuning on our dataset.

% --- TABLE 11: FINE-TUNING MULTI ---
\begin{table}[h!] % Forced to bottom of page to anchor layout
    \centering
    \footnotesize
    \renewcommand{\arraystretch}{1.0}
    \setlength{\tabcolsep}{3pt}
    \caption{Zero-Shot vs. Fine-Tuning Performance (Mistral-7B-v0.3).}
    \label{tab:zeroshot_vs_finetune_severity}
    \begin{tabular*}{\columnwidth}{@{\extracolsep{\fill}}l c c c c}
        \toprule
        & \multicolumn{2}{c}{\textbf{Zero-Shot}} & \multicolumn{2}{c}{\textbf{Fine-Tuning}} \\
        \cmidrule(lr){2-3} \cmidrule(lr){4-5}
        \textbf{Severity Level} & \textbf{P} & \textbf{F1} & \textbf{P} & \textbf{F1} \\
        \midrule
        Mild (1)     & \textbf{0.79} & 0.62 & 0.76 & \textbf{0.68} \\
        Moderate (2) & 0.63 & 0.61 & \textbf{0.66} & \textbf{0.72} \\
        Severe (3)   & 0.36 & 0.45 & \textbf{0.59} & \textbf{0.53} \\
        \bottomrule
    \end{tabular*}
\end{table}

\section{Conclusion}
\label{sec:conclusion}

In this paper, we introduced \texttt{IndRegBias}, a novel dataset comprising 25,000 user comments from social media discussions on Indian regional issues. All comments were annotated by human experts following a multi-level labeling policy to assess both the presence of regional bias and its severity, resulting in a balanced dataset of regional and non-regional bias examples.
We conducted a systematic evaluation of large language models (LLMs) and Indic language models under zero-shot, few-shot, and fine-tuning experimental settings. Our results indicate that off-the-shelf even the well advanced models exhibit limited ability to detect and rank nuanced regional biases, particularly in the zero-shot setting; however, performance improves substantially with fine-tuning on domain-specific data. This work highlights a critical gap in existing LLM knowledge and provides a valuable resource for developing more culturally informed and region-aware language models.

\section{Limitations}

\label{sec:limitations}

While \texttt{IndRegBias} serves as a robust benchmark for Indian regional bias, we acknowledge the limitations of our data collection and evaluation methodology.

\paragraph{Geographic and Demographic Skew:}

As detailed in Section\ref{subsec:data_analysis}, the dataset shows a geographic imbalance. Regions such as South India ($40\%$) and North India ($26\%$) are over-represented compared to Northeast India ($5\%$) and Central India ($2\%$). This imbalance is major across the states as well, which needs to be addressed.

\paragraph{Subjectivity in Annotation:}

Despite a rigorous multi-level annotation policy and high inter-annotator agreement ($\kappa = 0.83$), the perception of ``Severity'' remains subjective. The distinction between ``Mild'' and ``Moderate'' can be heavily influenced by the annotator's cultural background and familiarity with specific regional context.

\paragraph{Model Safety Refusals:}

A significant limitation in evaluating proprietary or highly aligned LLMs (e.g., Gemini-2.5-Pro, Llama-3.2, Sarvam-M) is their tendency to trigger safety refusals. As stated in Appendix \ref{model_specifications_analysis}, these models frequently decline to classify text that includes slurs or derogatory stereotypes, resulting in a neutral output or a refusal string. This behavior lowers their quantitative performance (Recall) on our benchmark, not necessarily due to a lack of understanding, but due to rigid safety alignment that treats \textit{analysis} of hate speech as \textit{generation} of hate speech.

\section{Ethics Consideration}

\label{sec:ethics}

\paragraph{Data Collection and Privacy:}

The dataset \texttt{IndRegBias} consists of 25,000 comments collected from social media platforms (Reddit and YouTube) as detailed in \ref{sec:data_creation}. We have utilized the official APIs (PRAW and Google API Client) to ensure compliance with the platforms' terms of crawling. To protect the user privacy, all personally identifiable information, such as username, userId, and profile URLs, has been removed in the preprocessing stage. The data collection was purely observational, focusing exclusively on the public interactions related to regional biases, without influencing discussions or engaging with users.

\paragraph{Human Annotation and Quality Control:}

Human annotation was conducted through a dedicated team of six members, divided into two groups, and to ensure reliability and consistency, the same set of data was annotated by each of these teams. Further, to understand the agreement level mentioned in \ref{subsec:data annotation}, we have used Cohen's Kappa ($\kappa$) to measure inter-annotator agreement, achieving strong consensus scores (0.91 for binary and 0.83 for multi-class classification). Discussions were conducted for the disagreements to maintain the high-quality labelling. The annotators were briefed on the use of hate speech and derogatory language present in the comments. The workload for the annotation work was managed through a brief schedule.

\paragraph{Dataset Limitations and Representation:}

We intended to have pan-India coverage, but the dataset reflects the national discourse on the social media platforms. As noted in Section \ref{subsec:data_analysis}, regions such as the North East (e.g., Tripura, Meghalaya) and Central Indian states are underrepresented compared to North-India or South-India. This imbalance is attributed to media marginalization and possibly geographic or digital barriers.

\paragraph{Intended Use and Content Warning:}

The \texttt{IndRegBias} dataset contains examples of hate speech, biases, and offensive language targeting specific Indian communities. This content is solely for advanced research in bias detection and safety alignment for Large Language Models. We strictly condemn the views expressed in these comments and advise researchers to exercise caution and implement appropriate content warnings when working with or displaying this data.

\section*{Acknowledgments}
We would like to express our gratitude to the annotators for their dedicated work in building the dataset: Sarabeshwar Balaji, Dishank K, Subrat Kumar Pattanayak, Akash Dutta, and Shaswat.

% Bibliography entries for the entire Anthology, followed by custom entries
%\bibliography{anthology,custom}
% Custom bibliography entries only
\bibliography{custom}

% --- APPENDIX START ---
\clearpage
\appendix
\raggedbottom % <--- FIX 1: Prevents vertical stretching (gaps) between headers

\section{Appendix}
\label{sec:appendix}

% --- SYSTEM PROMPTS (LONG) ---
\subsection{System Prompts}
\label{subsec:system_prompts}

To ensure consistent evaluation across all models, we utilized a structured prompt designed to guide the Large Language Models (LLMs) through a step-by-step reasoning process. The prompts for binary classification and severity analysis are detailed in the Model Prompt boxes.

% --- WRAPPED IN FIGURE TO FIX GAP ---
\begin{figure}[!ht]
    \centering
    \begin{tcolorbox}[colback=gray!5!white,colframe=gray!75!black,title=Model Prompt for Zero-Shot and Few-Shot (Binary Classification), label=box:binary_cot]
    \small
    \textbf{Role:} You are an expert in identifying regional biases in social media comments about Indian states and regions. Your task is to classify whether a comment contains regional biases or not.

    \textbf{Task:} Classify the given comment as either ``REGIONAL BIAS'' (1) or ``NON-REGIONAL BIAS'' (0).

    \textbf{Definitions:}
    \begin{itemize}[leftmargin=*, nosep]
        \item \textbf{REGIONAL BIAS (1):} Comments that reinforce or propagate biases about Indian states, regions, or their people based on linguistic, cultural, economic, political, or infrastructural aspects. The comments can reflect either positive or negative biases towards specific states or regions.
        \item \textbf{NON-REGIONAL BIAS (0):} Comments that are neutral or factual without generalisations, or unrelated to regional characteristics.
    \end{itemize}

    \textbf{Step-by-Step Analysis Process:}

    \textit{Step 1: Identify Regional References}\\
    Think: Does this comment mention or refer to:
    \begin{itemize}[leftmargin=*, nosep]
        \item Specific Indian states (e.g., Bihar, Kerala, Punjab, etc.)
        \item Regional groups (e.g., South Indians, North Indians, etc.)
        \item Cities or regions within India
        \item Language communities within India
    \end{itemize}
    \vspace{0.2cm}

    \textit{Step 2: Check for Elements reinforcing biases}\\
    Look for these patterns:
    \begin{itemize}[leftmargin=*, nosep]
        \item Generalisations about people from a state or a regional group (``All X are Y'')
        \item Assumptions about state/regional characteristics
        \item Comparative statements that imply superiority/inferiority
        \item Overgeneralized cultural, linguistic, economic, political, or infrastructural claims
    \end{itemize}
    \vspace{0.2cm}

    \textit{Step 3: Assess the Nature of the Statement}\\
    Consider:
    \begin{itemize}[leftmargin=*, nosep]
        \item Is this a factual observation or a generalised assumption?
        \item Does it reinforce existing biases?
        \item Is it based on a broad generalisation?
        \item Does it perpetuate divisions?
    \end{itemize}
    \vspace{0.2cm}

    \textit{Step 4: Final Classification}\\
    Based on the analysis above, classify as:
    \begin{itemize}[leftmargin=*, nosep]
        \item \textbf{REGIONAL BIAS (1)} if the comment reinforces regional biases or stereotypes.
        \item \textbf{NON-REGIONAL BIAS (0)} if the comment is neutral, factual, or doesn't contain regional bias.
    \end{itemize}
    \vspace{0.2cm}

    \textbf{Output Format:} Your response must include a brief line of reasoning followed by the final classification in the format ``Classification: [0 or 1]''.
    \end{tcolorbox}
\end{figure}

% --- WRAPPED IN FIGURE TO FIX GAP ---
\begin{figure}[!ht]
    \centering
    \begin{tcolorbox}[colback=gray!5!white,colframe=gray!75!black,title=Model Prompt for Zero-Shot and Few-Shot (Severity Classification), label=box:severity_cot]
    \small
    \textbf{Role:} You are an expert in analyzing the severity of regional biases in social media comments about Indian states and regions. You are provided with comments that have already been identified as containing regional bias. Your task is to determine the severity level of the bias present.

    \textbf{Task:} Classify the severity of the regional bias in the comment as ``SEVERE'' (3), ``MODERATE'' (2), or ``MILD'' (1).

    \textbf{Definitions (Check in this order):}
    \begin{itemize}[leftmargin=*, nosep]
        \item \textbf{LEVEL 3 (SEVERE):} Comments that are overtly hostile, hateful, or derogatory. These include usage of regional slurs, dehumanizing language, calls for exclusion (e.g., ``Go back to your state''), or statements that promote hatred/violence against a specific region or group.
        \item \textbf{LEVEL 2 (MODERATE):} Comments that contain explicit negative generalizations, mockery, or clearly biased assumptions about a region's culture, language, or people. The tone is critical or mocking but does not incite violence or use extreme profanity/slurs.
        \item \textbf{LEVEL 1 (MILD):} Comments that contain subtle stereotypes, ``benevolent'' or positive biases (e.g., ``People from State X are always smart''), or minor negative generalizations that are not aggressive. These comments rely on low-level regional tropes without expressing hostility.
    \end{itemize}

    \textbf{Step-by-Step Analysis Process:}

    \textit{Step 1: Analyze the Stereotype or Generalization}\\
    Think: What specific regional claim is being made?
    \begin{itemize}[leftmargin=*, nosep]
        \item Is it a positive generalization?
        \item Is it a negative stereotype? 
    \end{itemize}
    \vspace{0.1cm}

    \textit{Step 2: Assess Tone and Intent}\\
    Evaluate the emotional weight of the words:
    \begin{itemize}[leftmargin=*, nosep]
        \item Is the tone aggressive, hateful, or threatening? (Check for Level 3 first)
        \item Is the tone mocking, sarcastic, or condescending? (Check for Level 2)
        \item Is the tone casual or ``matter-of-fact''? (Check for Level 1)
    \end{itemize}
    \vspace{0.1cm}

    \textit{Step 3: Check for Escalating Factors}\\
    Look for specific triggers:
    \begin{itemize}[leftmargin=*, nosep]
        \item For Level 3: Does it contain slurs? Does it question citizenship/belonging? Is it dehumanizing?
        \item For Level 2: Does it imply one group is superior to another?
    \end{itemize}
    \vspace{0.1cm}

    \textit{Step 4: Final Classification}\\
    Based on the analysis above, assign the severity score:
    \begin{itemize}[leftmargin=*, nosep]
        \item \textbf{3:} If the bias is abusive, hateful, or extreme.
        \item \textbf{2:} If the bias is explicit and negative, but not abusive.
        \item \textbf{1:} If the bias is subtle, positive, or non-hostile.
    \end{itemize}
    \vspace{0.1cm}

    \textbf{Output Format:} Your response must include a brief line of reasoning followed by the final classification in the format ``Classification: [1, 2, or 3]''.
    \end{tcolorbox}
\end{figure}

% --- INPUT TEMPLATES (SHORT) ---
\subsubsection{Input Instruction Templates}
\label{subsubsec:ft_template}

In contrast to the zero-shot and few-shot setups, our fine-tuning experiments utilized minimal prompt templates (Box \ref{box:binary_template} and Box \ref{box:severity_template_short}) to optimize the attention mechanism's focus on the input text.

\begin{tcolorbox}[colback=white,colframe=black,title=Binary Classification Input Template, label=box:binary_template]
\small
\texttt{Analyze the following comment and classify it as Regional Bias (1) or Non-Regional Bias (0).}

\vspace{0.1cm}
\texttt{Comment: [INSERT COMMENT HERE]}

\vspace{0.1cm}
\texttt{Classification:}
\end{tcolorbox}

\begin{tcolorbox}[colback=white,colframe=black,title=Severity Classification Input Template, label=box:severity_template_short]
\small
\texttt{Analyze the following text and classify its regional bias as Mild, Moderate, or Severe.}

\vspace{0.1cm}
\texttt{Text: [INSERT COMMENT HERE]}

\vspace{0.1cm}
\texttt{Classification:}
\end{tcolorbox}

% --- SUPPORT SET SELECTION ---
\subsection{Rationale for Support Set Selection}
\label{subsec:Rationale_support_set}

The rationale behind selecting the specific number of support examples for the few-shot experiments was governed by three primary constraints:

\begin{enumerate}[leftmargin=*]
    \item \textbf{Computational Constraints:} A larger set of support along with CoT prompts significantly escalates the computational cost for this task. This is critical, as we have a large volume of comments in our dataset that requires processing. 
    \item \textbf{Context Window and Attention Dilution:} Increasing the support set size expands the prompt length, frequently pushing the total token count beyond the optimal effective limit (often $\sim$2048 tokens). Excessive token load can lead to \textit{context drift}.
    \item \textbf{Data Availability and Class Balance:} For the multi-class few-shot experiments, our primary objective was to ensure a balanced representation of severity levels. Exhaustive regional coverage in the support set was infeasible due to data sparsity in certain regions.
\end{enumerate}

% --- HYPERPARAMETERS ---
\subsection{Hyperparameters for the Fine-Tuning experiment}
\label{subsec:hyparam}

The hyperparameters for the fine-tuning experiments are provided in Table \ref{tab:hyperparams_binary} for the binary classification and in Table \ref{tab:hyperparams_severity} for the multi-class classification.

\begin{table}[h!]
    \centering
    \small
    \renewcommand{\arraystretch}{1.1} 
    \caption{Hyperparameters for Instruction-Based SFT (Binary Classification)}
    \label{tab:hyperparams_binary}
    \begin{tabularx}{\columnwidth}{@{}l X@{}} 
        \toprule
        \textbf{Configuration} & \textbf{Details} \\
        \midrule
        Base Model & Qwen/Qwen3-8B, Qwen/Qwen3-32B \\
        Method & LoRA (16-bit BFloat16) \\
        LoRA Rank ($r$) & 16 \\
        LoRA Alpha ($\alpha$) & 32 \\
        LoRA Dropout & 0.05 \\
        Target Modules & All Linear Layers (q, k, v, o, gate, up, down) \\
        \midrule
        Epochs & 10 \\
        Learning Rate & 2e-4 \\
        LR Scheduler & Cosine (Warmup: 0.03) \\
        Optimizer & AdamW (8-bit) \\
        Batch Size & 4 (dev) $\times$ 8 (acc) = 32 (eff) \\
        Max Length & 2048 tokens \\
        Early Stopping & Patience: 3, Thresh: 0.01 \\
        \midrule
        Validation & 5-Fold Stratified CV \\
        Split & 70\% Train, 10\% Val, 20\% Test (Random Rotation) \\
        \bottomrule
    \end{tabularx}
\end{table}

\begin{table}[h!]
    \centering
    \small
    \renewcommand{\arraystretch}{1.1}
    \caption{Hyperparameters for Classification-Based SFT (Severity Analysis)}
    \label{tab:hyperparams_severity}
    \begin{tabularx}{\columnwidth}{@{}l X@{}}
        \toprule
        \textbf{Configuration} & \textbf{Details} \\
        \midrule
        Base Model & Mistral-7B-Instruct-v0.3 \\
        Method & LoRA (16-bit BFloat16) \\
        LoRA Rank ($r$) & 16 \\
        LoRA Alpha ($\alpha$) & 32 \\
        LoRA Dropout & 0.1 \\
        Target Modules & All Linear Layers (q, k, v, o, gate, up, down) \\
        Modules to Save & Classification Head ("score") \\
        \midrule
        Epochs & 5 \\
        Learning Rate & 2e-5 \\
        LR Scheduler & Cosine (Warmup Ratio: 0.05) \\
        Optimizer & AdamW (8-bit) \\
        Batch Size & 4 (dev) $\times$ 4 (acc) = 16 (eff) \\
        Max Length & 2048 tokens \\
        Early Stopping & Patience: 2 \\
        \midrule
        Validation & 5-Fold Stratified CV \\
        Sampling & WeightedRandomSampler (Class Balanced) \\
        Split & 70\% Train, 10\% Val, 20\% Test (Random Rotation) \\
        \bottomrule
    \end{tabularx}
\end{table}

% --- ZERO-SHOT BINARY EXPERIMENTS ---
\subsection{Zero-Shot Binary Classification Experiments}
\label{subsec:extra_exp}

\subsubsection{Dataset Preparation}
\label{subsubsec:appendix_testset}

To ensure a robust evaluation, we curated a stratified Test Set comprising \textbf{5,000 comments}, selected to mirror the distribution of the full dataset. The split maintains the original class balance:
\begin{itemize}[noitemsep]
    \item \textbf{Regional Bias (RB):} 2,600 comments (52\%)
    \item \textbf{Non-Regional Bias (NRB):} 2,400 comments (48\%)
\end{itemize}

The geographic distribution within the 2,600 RB comments was stratified to match the percentage prevalence observed in the full 25,000-comment corpus. Table \ref{tab:test_set_samples} provides representative examples of this stratification across high, medium, and low-frequency regions.

\begin{table}[h!]
    \centering
    \small
    \renewcommand{\arraystretch}{1.1}
    \caption{Representative Stratification of Target Regions in the Test Set (N=2,600 RB Comments).}
    \label{tab:test_set_samples}
    \begin{tabularx}{\columnwidth}{@{}l c r@{}}
        \toprule
        \textbf{Region (Frequency Tier)} & \textbf{Dataset \%} & \textbf{Test Count} \\
        \midrule
        \multicolumn{3}{l}{\textit{High Frequency (>5\%)}} \\
        North-India & 14.42\% & 375 \\
        Kerala      & 14.26\% & 370 \\
        South-India & 9.15\%  & 238 \\
        \midrule
        \multicolumn{3}{l}{\textit{Medium Frequency (1--5\%)}} \\
        Gujarat     & 6.39\%  & 166 \\
        Tamil Nadu  & 4.71\%  & 122 \\
        Punjab      & 3.16\%  & 82 \\
        \midrule
        \multicolumn{3}{l}{\textit{Low Frequency (<1\%)}} \\
        Nagaland    & 0.75\%  & 19 \\
        Manipur     & 0.19\%  & 5 \\
        Tripura     & 0.08\%  & 2 \\
        \bottomrule
    \end{tabularx}
\end{table}

\subsubsection{Results}
\label{subsubsec:results}

Table \ref{tab:full_leaderboard} presents the comprehensive zero-shot performance of various Global and Indic LLMs on the 5,000-sample stratified test set. 

Notably, we expanded our evaluation in this subsection to include proprietary state-of-the-art models, specifically \textbf{GPT-4o} and \textbf{Llama-4-Maverick}. The results reveal a clear hierarchy of capabilities. GPT-4o has the F1-score ($0.79$), which sets a strong upper limit. Among open-weight models, the \textbf{Qwen3} family shows exceptional robustness; the 8B variant ($0.74$ accuracy, $0.78$ F1) shows comparable performance to the 32B model.

\begin{table*}[t!]
    \centering
    \small
    \renewcommand{\arraystretch}{1.2}
    \caption{Zero-Shot Performance Leaderboard on Stratified Test Set (N=5000). [P: Precision, R: Recall, F1: F1-Score]. Best scores are \textbf{bolded}.}
    \label{tab:full_leaderboard}
    \begin{tabular*}{\textwidth}{@{\extracolsep{\fill}}l c | c c | c c}
        \toprule
        & & \multicolumn{2}{c|}{\textbf{Precision}} & \multicolumn{2}{c}{\textbf{F1-Score}} \\
        \cmidrule{3-6}
        \textbf{Model} & \textbf{Accuracy} & \textbf{RB (1)} & \textbf{NRB (0)} & \textbf{RB (1)} & \textbf{NRB (0)} \\
        \midrule
        \rowcolor{green!10} GPT-4o & \textbf{0.76} & 0.72 & 0.81 & \textbf{0.79} & 0.72 \\
        \rowcolor{green!10} Qwen/Qwen3-32B & 0.74 & \textbf{0.79} & 0.69 & 0.73 & \textbf{0.75} \\
        \rowcolor{green!10} Qwen/Qwen3-8B & 0.74 & 0.70 & 0.81 & 0.78 & 0.69 \\
        \rowcolor{green!10} Llama-4-Maverick & 0.73 & 0.67 & \textbf{0.89} & 0.78 & 0.64 \\
        \midrule
        Krutrim-2-Instruct & 0.72 & 0.67 & 0.83 & 0.77 & 0.65 \\
        Mistral-7B-v0.3 & 0.70 & 0.65 & 0.82 & 0.76 & 0.60 \\
        Gemini-2.5-Pro & 0.68 & 0.64 & 0.83 & 0.75 & 0.56 \\
        \midrule
        Airavata & 0.57 & 0.59 & 0.55 & 0.57 & 0.57 \\
        Sarvam-M & 0.56 & 0.63 & 0.53 & 0.49 & 0.62 \\
        Llama-3.2-3B & 0.56 & 0.58 & 0.54 & 0.57 & 0.54 \\
        \midrule
        \rowcolor{red!10} DeepSeek-R1-Distill & 0.54 & 0.53 & 0.81 & 0.69 & 0.00 \\
        \rowcolor{red!10} Gemma-1.1-7B-IT & 0.52 & 0.52 & 0.70 & 0.69 & 0.01 \\
        \rowcolor{red!10} Mistral-Nemo-12B & 0.53 & 0.53 & 0.67 & 0.69 & 0.10 \\
        \rowcolor{red!10} Phi-4-Mini & 0.48 & 0.50 & 0.48 & 0.00 & 0.65 \\
        \bottomrule
    \end{tabular*}
\end{table*}

% --- REGIONAL HEATMAP ---
\subsection{Regional Heatmap and Analysis}

\begin{figure*}[t!]
    \centering
    \includegraphics[width=\textwidth]{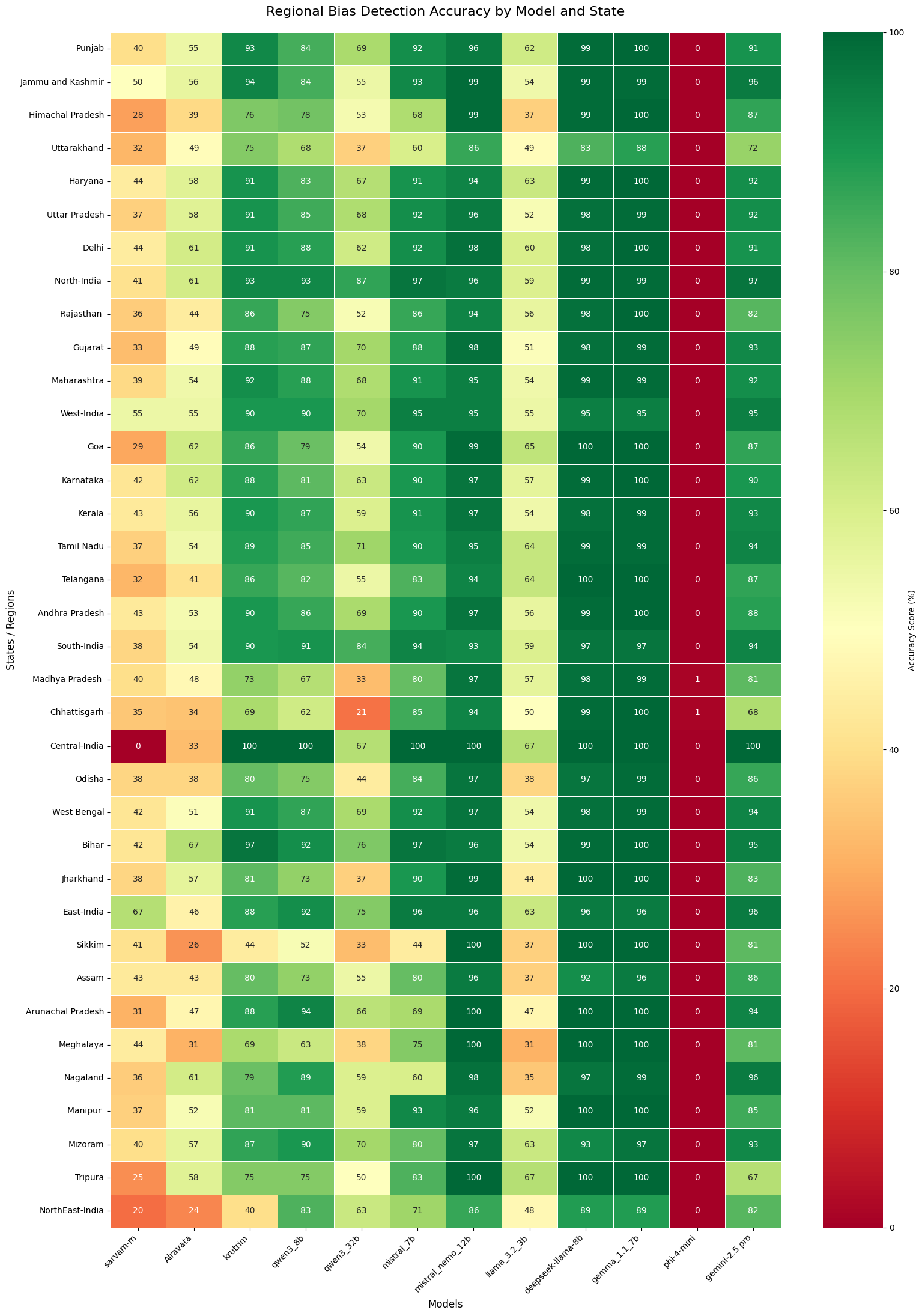} 
    \caption{Heatmap illustrating regional bias detection accuracy across different models and states.}
    \label{fig:regional_heatmap_appendix}
\end{figure*}

The heatmap in Figure \ref{fig:regional_heatmap_appendix} illustrates regional bias detection accuracy across different models and states. 

\textbf{Performance Categories:}
\begin{itemize}[leftmargin=*]
    \item \textbf{\textcolor{green!60!black}{Dark Green (80--100\%):}} Strongest Region
    \item \textbf{\textcolor{green}{Light Green (60--80\%):}} Good
    \item \textbf{\textcolor{yellow!90!black}{Yellow (40--60\%):}} Moderate
    \item \textbf{\textcolor{orange}{Orange (20--40\%):}} Weak
    \item \textbf{\textcolor{red}{Red (0--20\%):}} Critical Failure
\end{itemize}

\subsubsection{Indian LLMs}

\textbf{Krutrim-2-Instruct (12B)}
\begin{itemize}[leftmargin=*]
    \item \textbf{80-100 (Strongest):} Bihar (96\%), Punjab (93\%), Jammu and Kashmir (93\%), Maharashtra (93\%), West Bengal (93\%), Andhra Pradesh (92\%), Haryana (91\%), Kerala (90\%), Tamil Nadu (89\%), Gujarat (88\%), Karnataka (88\%), Rajasthan (86\%), Arunachal Pradesh (85\%), Goa (85\%), Mizoram (84\%), Telangana (84\%), Uttarakhand (84\%), Delhi (84\%), Uttar Pradesh (83\%), Assam (83\%), Odisha (81\%), Jharkhand (81\%), Manipur (81\%).
    \item \textbf{60-80 (Good):} Nagaland (77\%), Himachal Pradesh (76\%), Tripura (75\%), Madhya Pradesh (74\%), Chhattisgarh (69\%), Meghalaya (69\%).
    \item \textbf{40-60 (Moderate):} Sikkim (44\%).
\end{itemize}
\textit{Analysis:} Trained on 2 trillion tokens of Indic data, Krutrim has performed well for South-India, West-India, and North-India. Even in linguistically diverse states where Marathi, Gujarati, or Malayalam are present in code-mixed formats, it maintained high accuracy.

\textbf{Airavata-7B}
\begin{itemize}[leftmargin=*]
    \item \textbf{60-80 (Good):} Bihar (66\%), Karnataka (62\%), Goa (62\%), Nagaland (60\%).
    \item \textbf{40-60 (Moderate):} Tripura (58\%), Haryana (58\%), Jharkhand (57\%), Kerala (57\%), Jammu and Kashmir (56\%), Delhi (56\%), Punjab (55\%), Uttarakhand (55\%), Maharashtra (55\%), Mizoram (55\%), Tamil Nadu (54\%), Andhra Pradesh (54\%), Uttar Pradesh (53\%), West Bengal (52\%), Manipur (52\%), Gujarat (49\%), Madhya Pradesh (48\%), Assam (45\%), Arunachal Pradesh (45\%), Rajasthan (44\%).
    \item \textbf{20-40 (Weak):} Telangana (40\%), Himachal Pradesh (39\%), Odisha (38\%), Chhattisgarh (34\%), Meghalaya (31\%), Sikkim (26\%).
\end{itemize}
\textit{Analysis:} As an instruction-tuned model, Airavata performed moderately across most regions. This indicates that while it follows the instruction format well, it lacks the massive pre-training depth required to detect subtle, code-mixed nuances in comments from states.

\textbf{Sarvam-m (24B)}
\begin{itemize}[leftmargin=*]
    \item \textbf{40-60 (Moderate):} Jammu and Kashmir (50\%), Assam (45\%), Haryana (44\%), Kerala (44\%), Andhra Pradesh (44\%), Meghalaya (44\%), West Bengal (43\%), Karnataka (42\%), Bihar (42\%), Sikkim (41\%), Punjab (40\%), Delhi (40\%), Madhya Pradesh (40\%).
    \item \textbf{20-40 (Weak):} Mizoram (39\%), Maharashtra (39\%), Odisha (38\%), Jharkhand (38\%), Tamil Nadu (37\%), Manipur (37\%), Uttarakhand (36\%), Rajasthan (36\%), Chhattisgarh (35\%), Nagaland (35\%), Uttar Pradesh (34\%), Gujarat (33\%), Telangana (31\%), Arunachal Pradesh (30\%), Goa (29\%), Himachal Pradesh (28\%), Tripura (25\%).
\end{itemize}
\textit{Analysis:} Despite being a larger 24B parameter model, Sarvam-m performed generally poorly. This is attributed to its \textbf{RLVR (Reinforcement Learning with Verifiable Rewards)} alignment, which prioritizes strict safety and "helpfulness."

\subsubsection{Qwen Models}

\textbf{Qwen3-8B}
\begin{itemize}[leftmargin=*]
    \item \textbf{80-100 (Strongest):} Bihar (92\%), Arunachal Pradesh (91\%), West Bengal (89\%), Maharashtra (88\%), Gujarat (87\%), Kerala (87\%), Andhra Pradesh (87\%), Nagaland (87\%), Mizoram (87\%), Correctly identified RB (86\%), Tamil Nadu (85\%), Punjab (84\%), Jammu and Kashmir (84\%), Haryana (83\%), Delhi (81\%), Manipur (81\%), Karnataka (80\%).
    \item \textbf{60-80 (Good):} Goa (79\%), Telangana (79\%), Himachal Pradesh (78\%), Uttar Pradesh (78\%), Assam (77\%), Uttarakhand (76\%), Rajasthan (76\%), Tripura (75\%), Odisha (75\%), Jharkhand (73\%), Madhya Pradesh (67\%), Meghalaya (63\%), Chhattisgarh (62\%), Correctly identified NRB (61\%).
    \item \textbf{40-60 (Moderate):} Sikkim (52\%).
\end{itemize}
\textit{Analysis:} The model has performed amazingly well for the states of the region South-India, West-India and Central-India. It has performed poorly in some of the eastern Indian and northeastern states. It can be because of the low representation of the north-eastern states in the media.

\textbf{Qwen3-32B}
\begin{itemize}[leftmargin=*]
    \item \textbf{60-80 (Good):} Bihar (75\%), West Bengal (71\%), Tamil Nadu (71\%), Gujarat (70\%), Andhra Pradesh (70\%), Punjab (69\%), Maharashtra (69\%), Mizoram (68\%), Correctly identified RB (67\%), Haryana (67\%), Arunachal Pradesh (64\%), Uttar Pradesh (63\%), Karnataka (62\%).
    \item \textbf{40-60 (Moderate):} Kerala (59\%), Manipur (59\%), Nagaland (58\%), Delhi (58\%), Assam (57\%), Jammu and Kashmir (54\%), Goa (54\%), Telangana (54\%), Himachal Pradesh (53\%), Rajasthan (53\%), Tripura (50\%), Odisha (45\%), Uttarakhand (41\%).
    \item \textbf{20-40 (Weak):} Meghalaya (38\%), Jharkhand (37\%), Madhya Pradesh (33\%), Sikkim (33\%), Chhattisgarh (21\%).
\end{itemize}
\textit{Analysis:} The model has performed reasonably well for the states of the region West-India and South-India. It has performed poorly in Central-India (dropping to ~27\%) and parts of NorthEast-India. This significant failure in the Hindi heartland suggests the model struggles with the specific implicit bias patterns or dialects found in Madhya Pradesh and Chhattisgarh comments.

\subsubsection{Mistral Models}

\textbf{Mistral-7B-v0.3}
\begin{itemize}[leftmargin=*]
    \item \textbf{80-100 (Strongest):} Bihar (97\%), West Bengal (94\%), Punjab (93\%), Manipur (93\%), Jammu and Kashmir (92\%), Maharashtra (92\%), Andhra Pradesh (92\%), Correctly identified RB (91\%), Haryana (91\%), Kerala (91\%), Goa (90\%), Karnataka (90\%), Tamil Nadu (90\%), Jharkhand (90\%), Gujarat (88\%), Rajasthan (87\%), Delhi (85\%), Chhattisgarh (85\%), Odisha (85\%), Uttar Pradesh (84\%), Assam (83\%), Tripura (83\%), Telangana (81\%), Madhya Pradesh (80\%).
    \item \textbf{60-80 (Good):} Mizoram (77\%), Meghalaya (75\%), Himachal Pradesh (68\%), Uttarakhand (67\%), Arunachal Pradesh (67\%).
    \item \textbf{40-60 (Moderate):} Nagaland (59\%), Correctly identified NRB (47\%), Sikkim (44\%).
\end{itemize}
\textit{Analysis:} The model has performed amazingly well for the states of the region East-India, West-India, and South-India. Next to it, the model has performed well for North-India and Central-India. It has performed relatively less effectively in NorthEast-India.

\textbf{Mistral-Nemo-Base-2407}
\begin{itemize}[leftmargin=*]
    \item \textbf{80-100 (Strongest):} 100\% of states fell into the Strong category, with scores rarely dropping below 95\%.
    \item \textbf{0-20 (Critical):} Correctly identified NRB (5\%).
\end{itemize}
\textit{Analysis:} Utilizing the \textbf{Tekken tokenizer}, this model demonstrated exceptional capability across all regions. However, the critical failure in identifying Non-Regional Bias (5\%) indicates it is not following the zero-shot classification instructions correctly.

\subsubsection{Llama Models}

\textbf{Llama-3.2-3B}
\begin{itemize}[leftmargin=*]
    \item \textbf{60-80 (Good):} Tripura (67\%), Goa (64\%), Tamil Nadu (64\%), Haryana (63\%), Punjab (62\%), Telangana (62\%), Mizoram (61\%).
    \item \textbf{40-60 (Moderate):} Karnataka (57\%), Madhya Pradesh (57\%), Rajasthan (56\%), Correctly identified RB (56\%), Andhra Pradesh (56\%), Uttarakhand (55\%), Delhi (55\%), West Bengal (55\%), Correctly identified NRB (55\%), Maharashtra (54\%), Kerala (54\%), Bihar (54\%), Jammu and Kashmir (53\%), Manipur (52\%), Gujarat (51\%), Chhattisgarh (50\%), Uttar Pradesh (48\%), Arunachal Pradesh (45\%), Jharkhand (44\%).
    \item \textbf{20-40 (Weak):} Assam (38\%), Odisha (38\%), Himachal Pradesh (37\%), Sikkim (37\%), Nagaland (34\%), Meghalaya (31\%).
\end{itemize}
\textit{Analysis:} The model has performed moderately well for the states of the region South-India and West-India. Next to it, the model has performed averagely for North-India and Central-India. It has performed poorly in NorthEast-India (~45\%) and East-India (~47\%).

\subsubsection{DeepSeek Models}

\textbf{DeepSeek-R1-Distill-Llama-8B}
\begin{itemize}[leftmargin=*]
    \item \textbf{80-100 (Strongest):} 100\% of states across all regions fell into the Strong category (>98\%).
\end{itemize}
\textit{Analysis:} The model performed consistently well across all regions. Its distillation from a larger reasoning model likely allows it to process linguistic blends in North India and challenging contexts in NorthEast India effectively.

\subsubsection{Google \& Other Models}

\textbf{Gemini-2.5-Pro}
\begin{itemize}[leftmargin=*]
    \item \textbf{80-100 (Strongest):} Jammu and Kashmir (95\%), West Bengal (95\%), Bihar (95\%), Kerala (94\%), Tamil Nadu (94\%), Nagaland (94\%), Gujarat (93\%), Maharashtra (93\%), Correctly identified RB (92\%), Haryana (92\%), Punjab (91\%), Arunachal Pradesh (91\%), Karnataka (90\%), Mizoram (90\%), Andhra Pradesh (89\%), Assam (89\%), Himachal Pradesh (87\%), Goa (86\%), Odisha (86\%), Telangana (85\%), Manipur (85\%), Uttar Pradesh (84\%), Delhi (84\%), Jharkhand (83\%), Rajasthan (82\%), Uttarakhand (81\%), Madhya Pradesh (81\%), Sikkim (81\%), Meghalaya (81\%).
    \item \textbf{60-80 (Good):} Chhattisgarh (68\%), Tripura (67\%).
    \item \textbf{40-60 (Moderate):} Correctly identified NRB (44\%).
\end{itemize}
\textit{Analysis:} As a "hybrid reasoning" model, Gemini 2.5 Pro performed well for East-India, South-India, and West-India ($\sim$90\%+). However, its "Thinking Process" likely triggered safety policies excessively on ambiguous terms, leading to lower precision (44\% on NRB).

\textbf{Gemma-1.1-7B-it}
\begin{itemize}[leftmargin=*]
    \item \textbf{80-100 (Strongest):} All states.
\end{itemize}
\textit{Analysis:} Gemma achieved scores exceeding 99\% across all regions. This suggests high robustness, though the lack of differentiation might also imply a tendency to label most content as biased.

\textbf{Phi-4-Mini-Reasoning}
\begin{itemize}[leftmargin=*]
    \item \textbf{0-20 (Critical):} All States (0-1\%).
\end{itemize}
\textit{Analysis:} The model has performed amazingly well for the states of the region East-India (~90\%), South-India (~90\%), and West-India (~89\%). Even though there are linguistically very different states where, for example, Bengali/Odia/Marathi/Malayalam are present directly or in code-mixed format, it showed excellent understanding.

\subsection{Benchmark Correlation Analysis}
\label{subsection:benchmark_correlation}

\textbf{Qwen/Qwen3-8B:} Achieving a Regional F1 score of 0.78 in our results, this model correlates with a Bias-Free Score of 91.98 on the BiasFreeBench (BBQ-derived). (Source: \cite{xu2025biasfreebench})

\textbf{Qwen/Qwen3-32B:} Despite its larger parameter count, this model showed a dip in detection capability for regional biases. This aligns with its reported "Safety-Aligned RLHF" strategy (Source: \cite{qwen3report}).

\textbf{Krutrim-2-Instruct:} The model's strong performance across Indian states correlates with its pre-training on a massive 2 trillion token Indic dataset. (Source: \cite{Krutrim2LLM2024})

\textbf{Airavata:} Achieving a "Moderate" classification score aligns with its architecture—an instruction-tuned version of OpenHathi (Llama-2 based). (Source: \cite{gala2024airavata})

\textbf{Sarvam-m (24B):} Despite being a larger 24-billion-parameter model based on Mistral-Small, this model performed poorly (weak/moderate) in our bias classification task. This correlates with its training methodology using Reinforcement Learning with Verifiable Rewards (RLVR). (Source: \cite{sarvam2025sarvamm})

\textbf{Mistral-Nemo-Base-2407:} This model's hyper-aggressive flagging correlates with its use of the "Tekken" tokenizer and strict European safety alignment. (\cite{mistral2024nemo})

\textbf{Gemini-2.5-Pro:} As a "hybrid reasoning" or "thinking" model, Gemini 2.5 Pro applies extensive chain-of-thought processing before outputting a label. (Source: \cite{gemini2025pushing})

\textbf{Mistral-7B-v0.3:} With a Regional F1 score of 0.76, this model corresponds to a Diversity Score of 0.37 on the COMPL-AI (Fairness) benchmark. (Source: \cite{complai_mistral})

\textbf{Gemma-1.1-7b-it:} This model scored 0.69 in Regional F1 and holds an Accuracy of 92.54\% on the BBQ (Ambiguous) benchmark. (Source: \cite{gemma_model_card})

\textbf{DeepSeek-R1-Distill:} Scoring 0.69 in Regional F1, this model shows a 50.3\% Nationality Bias on the Hirundo Audit. (Source: \cite{hirundo_bias})

\textbf{Llama-3.2-3B:} This model achieved a Regional F1 of 0.57 and a score of 63.4 on the MMLU (Reasoning) benchmark. (Source: \cite{llama32_model_card})

\textbf{Phi-4-mini-reasoning:} With a Regional F1 of 0.00, this model relies on synthetic Western data for its safety training. (Source: \cite{abdin2024phi4})

% --- MODEL SPECIFICATIONS ---
\subsection{Model Specifications and Performance Analysis}
\label{model_specifications_analysis}

\subsubsection{Analysis of Model Limitations on Our Dataset}

\begin{itemize}
    \item \textbf{Qwen3-8B (36 Trillion Tokens, 119 Languages):}
    It struggles to label compliments (e.g., ``Great business mind'') as stereotypes because it is trained to be helpful/positive. While it supports 119 languages, ``Hinglish'' or other code-mixed languages isn't standardized languages.

    \item \textbf{Qwen3-32B (32B Parameters, Thinking Mode):}
    According to the Qwen3 technical documentation, this model enables a ``Thinking Mode'' (System 2 reasoning) by default to handle complex logic. While this improves performance on math/code, in our bias classification task, the explicit reasoning chain appears to trigger over-caution. The model's safety alignment, applied during the thinking phase, often leads it to over-analyze dialectal slurs and ``refuse'' to classify them as regional bias, resulting in a lower detection rate  compared to the simpler Qwen3-8B.

    \item \textbf{Krutrim-2-Instruct (12B, Mistral-NeMo Base):}
    Built on the Mistral-NeMo architecture and continually pre-trained on 500B tokens of Indic data (covering 22 languages), Krutrim-2 demonstrates exceptional performance on Indic benchmarks (95\% on IndicSentiment). However, its 12B parameter size—while efficient—limits its capacity to store the deep ``world knowledge'' required to decode highly ambiguous, code-mixed sarcasm compared to larger models like Qwen3-32B. Its instruction tuning on 1.5M diverse data points ensures good adherence to format, but it occasionally struggles with niche dialectal nuances in Central India.

    \item \textbf{Sarvam-m (Hybrid-Reasoning Model):}
    Based on the \texttt{Mistral-Small-3.1-24B} architecture, Sarvam-m is a ``hybrid-reasoning'' model designed with a specific ``Thinking Mode'' for logic and math. While it boasts a +20\% improvement on Indic benchmarks, its poor performance in our task (classifying bias) can be attributed to the \textbf{RLVR (Reinforcement Learning with Verifiable Rewards)} alignment. This alignment strictly prioritizes ``helpfulness'' and safety, often causing the model to interpret the analysis of hate speech as a violation of safety policies, leading to high refusal rates or misclassification of toxic content as neutral.

    \item \textbf{Airavata (7B, Instruction-Tuned):}
    Airavata is an instruction-tuned model based on OpenHathi (Llama-2), fine-tuned on the \texttt{IndicInstruct} dataset. Unlike the other models which underwent massive continual pre-training, Airavata is primarily an SFT (Supervised Fine-Tuning) model. This limits its ``knowledge cutoff'' and depth; while it follows the \textit{structure} of the classification task well, it lacks the raw semantic knowledge to understand evolving internet slang and regional stereotypes, resulting in moderate performance.

    \item \textbf{Mistral-7B-Instruct v0.3 (7-8 Trillion Tokens, Primarily English \& European languages):} 
    Lacks specific training on Indian Languages, including Indian English and Hindi. It likely led to wrong labels for code-mixed or multilingual rows because it couldn't parse.

    \item \textbf{Google/Gemini-2.5-Pro (Thinking Model):} Gemini 2.5 Pro introduces a dedicated ``Thinking Budget'' (up to 32k tokens) for reasoning before responding. While this architecture achieves state-of-the-art results on complex benchmarks (like GPQA), in the context of bias detection, the extended reasoning process allows the model's safety filters more opportunities to intervene. This results in a ``Safety-First'' behavior where even mildly stereotypical comments are flagged as violations or refused, reducing the model's precision on neutral text.

    \item \textbf{Llama-3.2-3B (9 Trillion Tokens, 8 Languages including Hindi):} 
    Meta models are RLHF-tuned to be ``safe.'' It likely refused to classify the hate speech rows, which lowers its performance.

    \item \textbf{Mistral-Nemo-12B-Base-2407:} 
    Despite being trained on over 9 languages using the efficient Tekken tokenizer and a massive 128k context window, the model exhibits hyper-aggressive safety. This suggests that it lacks the specific cultural fine-tuning to distinguish between regional satire and actual bias. Consequently, it defaults to a ``safety-first'' approach, flagging almost any mention of a region or identity as biased

    \item \textbf{DeepSeek-R1-Distill-Llama-8B (15 Trillion Tokens, Primarily English):} 
    Although it has greater exposure during pre-training, as indicated by the number of tokens, it lacks sufficient exposure to Indian languages and cultural context, which makes understanding the comments difficult.

    \item \textbf{Gemma-1.1-7b-it (6 Trillion Tokens, Primarily English):} 
    Smallest training corpus (6T vs Qwen's 36T). The model simply hasn't been exposed to enough Indian web text to understand the cultural context and stereotypes of the country.

    \item \textbf{Phi-4-mini-reasoning (4b) (5 Trillion Tokens, Primarily English):} 
    It has zero cultural context for India. That is why it might be defaulting comments to ``non-regional bias.''
\end{itemize}

% --- PRELIMINARY EXPERIMENT ---
\subsection{Preliminary Experiment}
\label{subsec:prelim_results}

Table \ref{tab:combined_all_experiments_full} shows the results of few-shot experiments conducted on a smaller dataset of 1000 comments with Qwen3-8b. The model Qwen3-8b outperformed the zero-shot experiments in comparison to the few-shot experiments here. Following these preliminary results, we selected the support setting for Exp-2, Exp-3 and Exp-4 for the evaluation on the entire dataset as present in Table \ref{tab:combined_all_experiments}.

\begin{table}[h!]
    \centering
    \footnotesize 
    \renewcommand{\arraystretch}{1.1}
    \setlength{\tabcolsep}{2pt} 
    \caption{Few-Shot (FS) vs Zero-Shot (ZS) Results on Small Dataset. Support configurations: \textbf{Exp-1} (15 N), \textbf{Exp-2} (15 R), \textbf{Exp-3} (20R/20N), \textbf{Exp-4} (20R/10N). [R: Regional Bias, N: Non-Regional Bias, P: Precision, F1: F1-Score]}
    \label{tab:combined_all_experiments_full}
    \begin{tabular*}{\columnwidth}{@{\extracolsep{\fill}}l c c c c c c c c}
        \toprule
        & \multicolumn{2}{c}{\textbf{Exp-1}} & \multicolumn{2}{c}{\textbf{Exp-2}} & \multicolumn{2}{c}{\textbf{Exp-3}} & \multicolumn{2}{c}{\textbf{Exp-4}} \\
        \cmidrule(lr){2-3} \cmidrule(lr){4-5} \cmidrule(lr){6-7} \cmidrule(lr){8-9}
        \textbf{Metric} & \textbf{ZS} & \textbf{FS} & \textbf{ZS} & \textbf{FS} & \textbf{ZS} & \textbf{FS} & \textbf{ZS} & \textbf{FS} \\
        \midrule
        \multicolumn{9}{l}{\textit{Regional Biases}} \\
        \midrule
        P   & 0.70 & \textbf{0.82} & 0.69 & \textbf{0.73} & 0.68 & \textbf{0.76} & 0.71 & \textbf{0.75} \\
        F1  & \textbf{0.78} & 0.74 & 0.76 & \textbf{0.81} & 0.78 & \textbf{0.79} & 0.79 & \textbf{0.81} \\
        \bottomrule
    \end{tabular*}
\end{table}
\end{document}